\documentclass[10pt,journal,compsoc]{IEEEtran}
\usepackage{float}
\usepackage{caption}
\usepackage{times}
\usepackage{graphicx}
\usepackage{amsmath}
\usepackage{amssymb}
\usepackage{mathrsfs}
\usepackage{booktabs}
\usepackage{makecell}
\usepackage{multirow}
\usepackage{algorithmic}
\usepackage{threeparttable}
\usepackage{colortbl}
\usepackage{tabu}
\usepackage{xspace}
\usepackage{stix}
\usepackage{pifont}
\usepackage{enumitem}
\usepackage{xcolor}
\usepackage{color}
\usepackage{cancel}
\usepackage{wasysym}
\usepackage{listings}
\usepackage{array}
\usepackage{booktabs}
\usepackage{algorithm}
\usepackage{makecell}
\usepackage{wasysym}
\usepackage{ifsym}
\usepackage{ragged2e}

\definecolor{myorange}{RGB}{255,100,3}
\definecolor{mygray}{gray}{.85}
\definecolor{mygray1}{gray}{.7}
\definecolor{mygray2}{gray}{.93}
\definecolor{mygray3}{gray}{.90}

\definecolor{votblue}{RGB}{180, 220, 255}        
\definecolor{votbluelight}{RGB}{230, 240, 250}   

\definecolor{votgreen}{RGB}{180, 240, 200}       
\definecolor{votgreenlight}{RGB}{230, 245, 235}  

\definecolor{votorange}{RGB}{255, 210, 160}      
\definecolor{votorangelight}{RGB}{255, 235, 220} 

\usepackage{amssymb,graphicx}
\newcommand\vcent[1]{\vcenter{\hbox{#1}}}
\newcommand\loudspeaker[1][3]{\ensuremath{\vcent{\rule{.6ex}{.6ex}}\kern-.5ex%
  \vcent{\scalebox{.6}[1]{\rotatebox[origin=center]{90}{$\blacktriangle$}}}%
  \ifnum#1>0\relax\kern.1ex\vcent{\scalebox{.3}{)}}\ifnum#1>1\relax\kern-.15ex%
  \vcent{\scalebox{.4}{)}}\ifnum#1>2\relax\kern-.23ex\vcent{\scalebox{.5}{)}}%
  \fi\fi\fi}%
}

\usepackage{epsfig}
\newcommand{\pub}[1]{\color{gray}{\scriptsize{[{#1}]}}}

\usepackage{mathrsfs}
\usepackage{soul}
\newcommand{\ie}{{\emph{i.e.}}\xspace}
\newcommand{\eg}{{\emph{e.g.}}\xspace}

\newcommand{\vs}{{\emph{vs.}}\xspace}
\newcommand{\etal}{{\emph{et al.}}\xspace}
\newcommand{\etc}{{\emph{etc.}}\xspace}
\usepackage{array}
\usepackage{tabu}

\usepackage[pagebackref,breaklinks,colorlinks]{hyperref}

\usepackage[capitalize]{cleveref}
\crefname{section}{Sec.}{Secs.}
\Crefname{section}{Section}{Sections}
\Crefname{TABLE}{Table}{Tables}
\crefname{table}{TABLE}{TABLE}

\newcommand{\Rmnum}[1]{\expandafter\@slowromancap\romannumeral #1@}

\usepackage{color}
\usepackage{xcolor}
\newcommand{\numvideo}{5,024\xspace}
\newcommand{\numobject}{10,074\xspace}
\newcommand{\numclass}{200\xspace}
\newcommand{\nummask}{701,976\xspace}
\newcommand{\numframes}{468,251\xspace}
\newcommand{\numduration}{1,570.63\xspace}
\newcommand{\numbaseline}{20\xspace}

\newcommand{\ournewdataset}[1]{MOSEv#1\xspace}
\newcommand{\J}{$\mathcal{J}$\xspace}
\newcommand{\F}{$\mathcal{F}$\xspace}
\newcommand{\JF}{$\mathcal{J}\&\mathcal{F}$\xspace}

\newcommand{\JFnd}{$\mathcal{J}\&\dot{\mathcal{F}}_d$\xspace}
\newcommand{\JFnr}{$\mathcal{J}\&\dot{\mathcal{F}}_r$\xspace}
\newcommand{\Fn}{$\dot{\mathcal{F}}$\xspace}
\newcommand{\JFn}{$\mathcal{J}\&\dot{\mathcal{F}}$\xspace}
\newcommand{\JFnbd}{$\mathcal{J}\&\dot{\mathcal{F}}_{bd}$\xspace}

\makeatletter
\newcommand{\thickhline}{%
    \noalign {\ifnum 0=`}\fi \hrule height 1pt
    \futurelet \reserved@a \@xhline
}
\makeatother

\makeatletter
\DeclareRobustCommand\onedot{\futurelet\@let@token\@onedot}
\def\@onedot{\ifx\@let@token.\else.\null\fi\xspace}

\def\eg{\emph{e.g}\onedot} 
\def\ie{\emph{i.e}\onedot} 
 
\def\etc{\emph{etc}\onedot} \def\vs{\emph{vs}\onedot}
 
\def\etal{\emph{et al}\onedot}

\definecolor{mycolor}{RGB}{0, 0, 0}

\usepackage{pifont}
\newcommand{\myparagraph}[1]{{\vspace{.16em} \noindent \bf #1}}

\let\oldsubsection\subsection
\renewcommand{\subsection}[1]{\oldsubsection{#1} }
\ifCLASSOPTIONcompsoc
  \usepackage[nocompress]{cite}
\else
  \usepackage{cite}
\fi
\ifCLASSINFOpdf
\else
\fi
\begin{document}
%
\title{MOSEv2: A More Challenging Dataset for \\ Video Object Segmentation in Complex Scenes}

\author{Henghui~Ding,
        Kaining~Ying,
        Chang~Liu,
        Shuting~He,
        Xudong~Jiang,~\IEEEmembership{Fellow,~IEEE},
        Yu-Gang~Jiang,~\IEEEmembership{Fellow,~IEEE},
        Philip~H.S.~Torr,
        Song~Bai
\IEEEcompsocitemizethanks{
\IEEEcompsocthanksitem Henghui Ding, Kaining Ying, and Yu-Gang Jiang are with Fudan University, Shanghai, China. (e-mail: henghui.ding@gmail.com)
\IEEEcompsocthanksitem Chang Liu and Song Bai are with ByteDance Inc.
\IEEEcompsocthanksitem Shuting He is with Shanghai University of Finance and Economics, China.
\IEEEcompsocthanksitem Xudong Jiang is with Nanyang Technological University, Singapore.
\IEEEcompsocthanksitem Philip H.S. Torr is with University of Oxford, United Kingdom.
\IEEEcompsocthanksitem Henghui Ding and Kaining Ying are co-first authors.
}
}

\markboth{MOSE Dataset Page: \href{https://MOSE.video}{MOSE.video}}%
{Shell \MakeLowercase{\textit{et al.}}: Bare Demo of IEEEtran.cls for Computer Society Journals}

\IEEEtitleabstractindextext{
\justify
\begin{abstract} 
   Video object segmentation (VOS) aims to segment specified target objects throughout a video. Although state-of-the-art methods have achieved impressive performance (\eg, 90+\% $\mathcal{J}\&\mathcal{F}$) on benchmarks such as DAVIS and YouTube-VOS, these datasets primarily contain salient, dominant, and isolated objects, limiting their generalization to real-world scenarios. To bridge this gap, the co\textbf{M}plex video \textbf{O}bject \textbf{SE}gmentation (\ournewdataset{1}) dataset was introduced to facilitate VOS research in complex scenes. Building on the foundations and insights of MOSEv1, we present \textbf{\ournewdataset{2}}, a significantly more challenging dataset designed to further advance VOS methods under real-world conditions. \ournewdataset{2} consists of \numvideo videos and \nummask high-quality masks for \numobject objects across \numclass categories. Compared to its predecessor, \ournewdataset{2} introduces much greater scene complexity, including {more frequent object disappearance and reappearance, severe occlusions and crowding, smaller objects, as well as a range of new challenges such as adverse weather (\eg, rain, snow, fog), low-light scenes (\eg, nighttime, underwater), multi-shot sequences, camouflaged objects, non-physical targets (\eg, shadows, reflections), and scenarios requiring external knowledge.} We benchmark \numbaseline representative VOS methods under 5 different settings and observe consistent performance drops on \ournewdataset{2}. For example, SAM2 drops from {76.4\%} on \ournewdataset{1} to only 50.9\% on \ournewdataset{2}. We further evaluate 9 video object tracking methods and observe similar declines, demonstrating that \ournewdataset{2} poses challenges across tasks. These results highlight that despite strong performance on existing datasets, current VOS methods still fall short under real-world complexities. Based on our analysis of the observed challenges, we further propose several practical tricks that enhance model performance. \ournewdataset{2} is publicly available at \href{https://MOSE.video}{https://MOSE.video}.
\end{abstract}

}

\maketitle


\IEEEdisplaynontitleabstractindextext

\IEEEpeerreviewmaketitle

\section{Introduction}
\label{sec:intro}

\begin{figure*}
\centering 
\includegraphics[width=0.9996\textwidth]{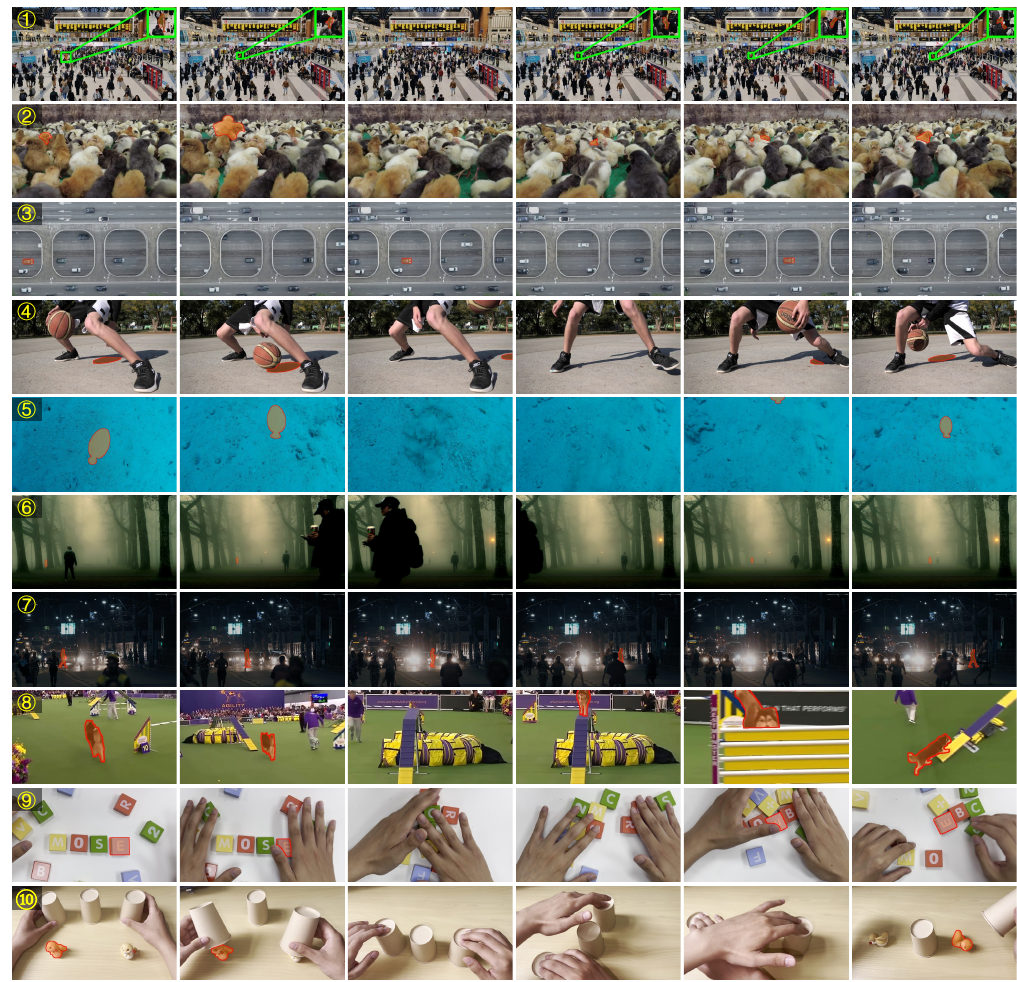} 
\vspace{-6.96mm}
\caption{
Example videos from the proposed \ournewdataset{2} dataset. 
Selected target objects are masked in \textcolor{myorange}{orange}. 
The target in case \raisebox{-0.1em}{\scalebox{1.06}{\ding{172}}} is enlarged for better visualization.
The most notable features of \ournewdataset{2} include both challenges inherited from \ournewdataset{1}~\cite{MOSEv1} such as object disappearance-reappearance (\raisebox{-0.1em}{\scalebox{1.06}{\ding{172}}}-\raisebox{-0.1em}{\scalebox{1.06}{\ding{181}}}), small/inconspicuous objects (\raisebox{-0.1em}{\scalebox{1.06}{\ding{172}}},\raisebox{-0.1em}{\scalebox{1.06}{\ding{174}}},\raisebox{-0.1em}{\scalebox{1.06}{\ding{177}}}), heavy occlusions (except \raisebox{-0.1em}{\scalebox{1.06}{\ding{176}}}), and crowded scenes (\raisebox{-0.1em}{\scalebox{1.06}{\ding{172}}},\raisebox{-0.1em}{\scalebox{1.06}{\ding{173}}}), as well as newly introduced complexities such as adverse weather (\raisebox{-0.1em}{\scalebox{1.06}{\ding{177}}}), low-light environments (\raisebox{-0.1em}{\scalebox{1.06}{\ding{176}}}-\raisebox{-0.1em}{\scalebox{1.06}{\ding{178}}}), multi-shots (\raisebox{-0.1em}{\scalebox{1.06}{\ding{179}}}), camouflaged objects (\raisebox{-0.1em}{\scalebox{1.06}{\ding{176}}}), non-physical objects (\raisebox{-0.1em}{\scalebox{1.06}{\ding{175}}}), and knowledge dependency (\raisebox{-0.1em}{\scalebox{1.06}{\ding{180}}},\raisebox{-0.1em}{\scalebox{1.06}{\ding{181}}}). The goal of \ournewdataset{2} dataset is to provide a platform that promotes the development of more comprehensive and robust video object segmentation algorithms.
}
\label{Fig:teaser}
\vspace{-3.36mm}
\end{figure*}

\IEEEPARstart{V}{ideo} object segmentation (VOS)~\cite{davis2016,davis2017,youtube_vos,MOSEv1} aims to segment specified target objects throughout an entire video. It is one of the most fundamental and challenging computer vision tasks, playing a crucial role in various practical applications involving video analysis and understanding, such as autonomous vehicle, augmented reality, and video editing. There are different settings for VOS, for example, semi-supervised VOS~\cite{DBLP:conf/cvpr/CaellesMPLCG17,Park_2021_CVPR} that gives the first-frame mask, bounding box, or points of the target object, unsupervised VOS~\cite{jain2017fusionseg,cheng2017segflow} that automatically finds primary or salient objects, and interactive VOS~\cite{chen2018blazingly,oh2019fast} that relies on user interactions with the target object. VOS has been extensively studied in the past using traditional techniques~\cite{brox2010object,lee2011key,JumpCut} and deep learning methods~\cite{cutie, SAM2}. Deep-learning-based approaches have greatly improved VOS performance and surpassed traditional techniques by a large margin.

Current state-of-the-art VOS methods~\cite{cutie,xmem,SAM2} have achieved near-saturation performance on two of the commonly-used VOS datasets DAVIS~\cite{davis2016, davis2017} and YouTube-VOS~\cite{youtube_vos}. For example, XMem~\cite{xmem} achieves {92.0\%} \JF on DAVIS~2016~\cite{davis2016}, {87.7\%} \JF on DAVIS~2017~\cite{davis2017}, and {86.1\%} \JF on YouTube-VOS~\cite{youtube_vos}. With such a high performance, it seems that video object segmentation has been well resolved. However, \textbf{\textit{do we really perceive objects in realistic scenarios}}?

To explore this question, we introduced the co\textbf{M}plex video \textbf{O}bject \textbf{SE}gmentation (\ournewdataset{1}) dataset in \cite{MOSEv1}, revisiting VOS under more realistic and complex scenes where traditional datasets fall short. In contrast to DAVIS~\cite{davis2016, davis2017} and YouTube-VOS~\cite{youtube_vos}, where target objects are typically salient and isolated, \ournewdataset{1} focuses on challenging cases such as object disappearance and reappearance, small or inconspicuous objects, heavy occlusions, and crowded scenes. These real-world conditions significantly affect segmentation performance, with XMem~\cite{xmem} only achieving 57.6\% \JF on \ournewdataset{1}. Since its release in 2023, \ournewdataset{1} has attracted broad and growing attention from the research community. Several competitions have been organized based on this dataset, including PVUW~\cite{ding2025pvuw,pvuw} and LSVOS~\cite{lsvos}, facilitating research in this area. Meanwhile, a series of strong VOS methods such as SAM2~\cite{SAM2} have subsequently pushed the performance from initial baselines to 76.4\% \JF, highlighting both the difficulty and the value of the dataset while demonstrating substantial progress in addressing complex video segmentation scenarios.

In this work, building on the foundations and insights of \ournewdataset{1}, we present \textbf{\ournewdataset{2}}, a more challenging dataset that further pushes the boundaries of VOS in real-world scenes. \ournewdataset{2} significantly increases the complexity across multiple dimensions. Core challenges of \ournewdataset{1}, such as object disappearance-reappearance, occlusions, small objects, and crowded scenes, are retained but appear more frequently, with greater severity, and under more realistic conditions. Beyond that, \ournewdataset{2} introduces a range of new challenges rarely covered in previous datasets, including adverse weather (\eg, rain, snow, fog), low-light scenes (\eg, nighttime, underwater), multi-shot sequences, camouflaged objects, non-physical targets (\eg, shadows, reflections), knowledge-dependent scenarios, \etc These additions aim to bridge the gap between current VOS datasets and the diverse, unconstrained nature of real-world scenes. With these multifaceted complexities, {\ournewdataset{2}} serves as a next-generation benchmark for evaluating and advancing complex video object segmentation under realistic, dynamic, and highly unconstrained environments.
\\
\hspace*{1.536em}
\ournewdataset{2} consists of \numvideo videos and \numobject annotated object instances spanning \numclass diverse categories, resulting in \nummask high-quality segmentation masks. 
Representative examples are shown in \cref{Fig:teaser}, illustrating both intensified and newly introduced challenges. A common pattern is object disappearance and reappearance, as shown in the 3rd example where a vehicle repeatedly disappears and reappears under overpasses, requiring robust temporal association. Challenges like small or inconspicuous objects, crowded scenes, and severe occlusions are also more prominent. For example, in the 1st example, a tiny person moves through a dense crowd, frequently occluded by others. Examples 4-10 highlight some new challenges in \ournewdataset{2}. Adverse weather (\eg, fog in the 6th), low-light conditions (\eg, underwater in the 5th, nighttime in the 7th), and multi-shot sequences (\eg, 8th) introduce appearance instability, motion ambiguity, and temporal discontinuities. These demand strong generalization and long-range association. Moreover, \ournewdataset{2} includes novel object categories that are difficult for existing methods. For example, camouflaged objects (5th) blend into backgrounds, while non-physical targets like shadows (4th) lack stable visual cues and change shape based on external factors. In addition, \ournewdataset{2} further introduces knowledge-dependent scenarios (\eg, 9th and 10th examples) that require high-level reasoning. For example, the 9th example requires optical character recognition to differentiate similar-looking blocks, while the 10th involves physics-based causality, where the target must be inferred from surrounding motion despite being invisible. These diverse and fine-grained challenges make \ournewdataset{2} a comprehensive dataset for studying the robustness and generalization capabilities of VOS in open-world complex scenes. We expect \ournewdataset{2} to spur meaningful progress toward real-world video understanding and deployment.

To thoroughly analyze the proposed \ournewdataset{2} dataset, we retrain and benchmark \numbaseline representative VOS methods under different settings. Experimental results demonstrate that the complexity of real-world videos in \ournewdataset{2} significantly degrades the performance of current state-of-the-art VOS methods. For example, the \JF score of SAM2~\cite{SAM2} reaches {90.7\%} on DAVIS~2017 \cite{davis2017} and {76.4\%} on \ournewdataset{1}~\cite{MOSEv1}, but notably drops to \textbf{50.9\%} on \ournewdataset{2}. Similarly, Cutie~\cite{cutie} achieves {87.9\%} on DAVIS~2017 and {69.9\%} on \ournewdataset{1}, but markedly declines to \textbf{43.9\%} on \ournewdataset{2}. These consistent performance drops highlight the significant challenges posed by the more realistic and complex scenarios in \ournewdataset{2}.

Beyond VOS, \ournewdataset{2} extends naturally to a wide range of video perception tasks requiring fine-grained understanding. In particular, we demonstrate its applicability to video object tracking (VOT) by benchmarking 9 state-of-the-art VOT methods~\cite{seqtrack,aqatrack,odtrack,lorat,sutrack,SAM2,SAMURAI,dam4sam,SAM2Long} on \ournewdataset{2}. While these methods perform well on standard VOT benchmarks such as LaSOT~\cite{lasot} and GOT-10k~\cite{got-10k}, consistent and notable performance drops are observed on \ournewdataset{2}. For example, SAMURAI~\cite{SAMURAI} achieves 74.2\% AUC on LaSOT but only 36.1\% on MOSEv2, revealing that \ournewdataset{2} introduces new and significant challenges not only for VOS but also for VOT. This demonstrates the broader applicability of MOSEv2 as a strong foundation for video understanding research in realistic and complex scenes.

In summary, our main contributions are as follows:
\begin{itemize}
\setlength\itemsep{0.36em}
    \item We present \ournewdataset{2} (co\textbf{M}plex video \textbf{O}bject \textbf{SE}gmentation), a more challenging dataset for video object segmentation in complex scenes. Compared to MOSEv1~\cite{MOSEv1}, \ournewdataset{2} introduces {more frequent object disappearance-reappearance, more severe occlusions, denser crowding, and smaller targets, and also new complexities such as adverse weather, low-light scenes, multi-shot videos, camouflaged objects, non-physical targets, and knowledge-dependent scenarios.}

    \item We provide detailed comparative analysis between \ournewdataset{2} and existing VOS and VOT datasets, highlighting its unique challenges and greater complexity that better represent real-world video understanding scenarios.
    
    \item We conduct comprehensive benchmarks of state-of-the-art methods on \ournewdataset{2} across various VOS and VOT settings, including semi-supervised VOS with mask, box, and point initialization, as well as unsupervised VOS, interactive VOS, and video object tracking.
    
    \item We perform an in-depth analysis of model performance and failure cases on \ournewdataset{2}, highlighting the key challenges it poses. Building on these insights, we propose practical tricks that substantially enhance model performance in complex scenarios, and outline future directions for advancing robust video understanding in the wild.
\end{itemize}
\section{Related Work}
\subsection{Video Object Segmentation}
Video object segmentation (VOS) aims to segment a specific object throughout a video. 
Based on how the target object is specified, VOS can be categorized into four main settings: 1) semi-supervised VOS (also known as semi-automatic VOS~\cite{VOS_Survey} or one-shot VOS), 2) unsupervised VOS (also called automatic VOS or zero-shot VOS), 3) interactive VOS, and 4) referring VOS.

\noindent$\bullet$~\textbf{Semi-supervised VOS.} Semi-supervised VOS~\cite{DBLP:conf/cvpr/CaellesMPLCG17} aims to segment the target object throughout a video, given its mask in the first frame. Most existing works can be categorized into propagation-based methods~\cite{DBLP:conf/cvpr/PerazziKBSS17,jang2017online,DBLP:conf/cvpr/JampaniGG17,xiao2018monet,hu2018motion,han2018reinforcement,youtube_vos,cheng2018fast,xu2019mhp,chen2020state,huang2020fast,wug2018fast,jabri2020space,lin2019agss,zhang2019fast} and matching-based methods~\cite{DBLP:conf/iccv/YoonRKLSK17,cheng2018fast,voigtlaender2019feelvos,wang2019ranet,Duarte_2019_ICCV,Oh_2019_ICCV,zhang2020transductive,Mast,yang2020collaborative,Hu_2021_CVPR,duke2021sstvos,cutie,xmem,xmem++}. Propagation-based methods leverage the predicted mask from the previous frame to guide the segmentation of the current frame, thereby propagating object cues in a frame-by-frame manner. Matching-based methods, on the other hand, first encode the target object into an embedding space and then perform per-pixel classification by comparing the similarity between each pixel’s feature and the stored object embedding.
Since obtaining pixel-level annotations is often expensive and time-consuming, some methods employ {bounding box} as the first-frame reference~\cite{SimMask, FTMU, lin2021query}. For example, SiamMask~\cite{SimMask} integrates a mask prediction branch into a fully convolutional Siamese object tracker to generate binary segmentation masks. 

 Recently, SAM2~\cite{SAM2} adopts promptable visual segmentation, which allows the model to accept prompts in the form of positive/negative clicks, bounding boxes, or masks on any frame of a video. This flexible interaction significantly improves the model’s adaptability and generalization across diverse scenarios. Following SAM2, several efficient extensions~\cite{SAM2Long, SAMURAI, dam4sam, yang2025mosam} have been proposed to improve its performance. For example, SAM2Long~\cite{SAM2Long} addresses error accumulation by exploring multiple segmentation pathways via constrained tree search. DAM4SAM~\cite{dam4sam} introduces a distractor-aware memory and an introspection-based update strategy to mitigate ambiguity from visual distractors. To better handle dynamic scenes, recent works~\cite{SAMURAI, yang2025mosam} incorporate motion modeling into promptable segmentation. SAMURAI~\cite{SAMURAI} integrates Kalman filtering~\cite{kalman1960new} for adaptive memory selection, while MoSAM~\cite{yang2025mosam} enhances robustness through motion-aware sparse and dense prompts combined with spatiotemporal memory mechanisms. These SAM2 variant methods achieve impressive performance on the previous VOS datasets~\cite{SAM2,youtube_vos,LVOSv1,LVOSv2,MOSEv1,davis2016,davis2017}.

\noindent$\bullet$~\textbf{Interactive VOS.} This task aims at segmenting the target object in a video indicated by user's interaction (\eg, clicks or scribbles) \cite{MiVOS,oh2019fast,GIS,MANet,chen2018blazingly,cheng2018fast,chen2020scribblebox,Yin_2021_CVPR}, it is a special form of semi-supervised VOS. Existing methods mainly follow a paradigm of interaction-propagation way. Besides the feature encoder that extracts pixel features, there are other two modules placed on the feature encoder to achieve interactive video object segmentation, \ie, interactive segmentation module that corrects prediction based on user's interaction and mask propagation module that propagates user-corrected masks to other frames. 
SAM2~\cite{SAM2} has also demonstrated strong capabilities in this task, offering superior performance with flexible interaction mechanisms, significantly enhancing both segmentation quality and user experience.

\noindent$\bullet$~\textbf{Referring VOS.} This is an emerging setting that aims to segment the target object in a video according to a text expression~\cite{MeViS, MeViSv2, seo2020urvos, ding2021vision,ye2021referring,move}. 
Early methods can be broadly classified as bottom-up methods and top-down methods.
Bottom-up methods~\cite{seo2020urvos, VLTPAMI, liu2021cmpc} perform first-frame segmentation followed by mask propagation or per-frame segmentation with post-hoc association. Top-down methods~\cite{Botach_2022_CVPR, Wu_2022_CVPR} first generate candidate tracklets and then select the one best aligned with the expression. The introduction of motion-centric datasets in MeViS~\cite{MeViS} and MeViSv2~\cite{MeViSv2} has drawn increased attention to the importance of temporal dynamics. Subsequent works~\cite{dshmp} highlight that temporal modeling is essential for accurate grounding. Recent works~\cite{visa,videolisa} also leverage multimodal large language models~\cite{llava,internvl} to handle expressions requiring complex reasoning, which enables human-like understanding and generalization ability across diverse language descriptions. With the latest datasets such as MeViSv2~\cite{MeViSv2} and OmniAVS~\cite{omniavs} supporting expressions across multiple modalities, omnimodal referring VOS is expected to gain increasing attention in future research~\cite{ReferringSurvey}.

\begin{figure*}[ht!]
    \centering
    \includegraphics[width=0.98\linewidth]{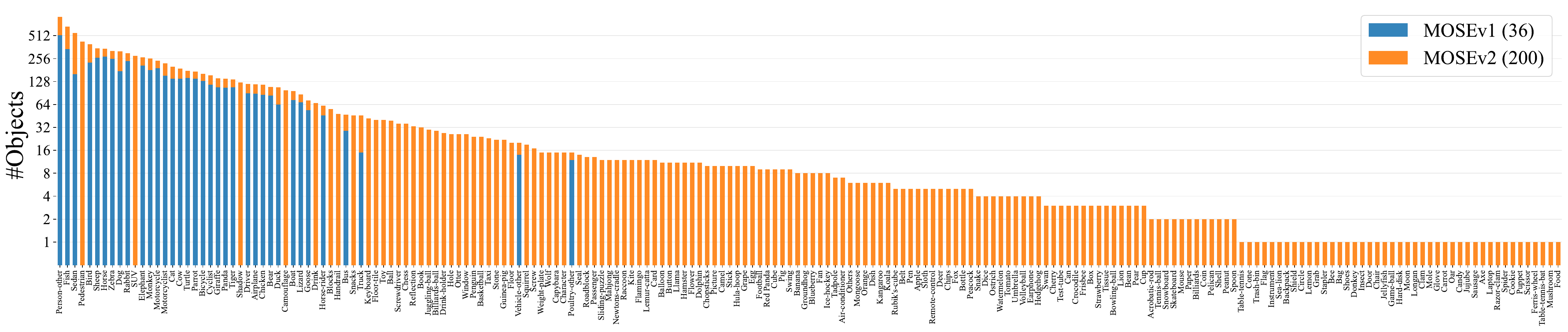}
    \vspace{-3mm}               
    \caption{Category distributions of \ournewdataset{1}~\cite{MOSEv1} and the proposed \ournewdataset{2}.}
    \label{fig:categories}
    \vspace{-2mm}
\end{figure*}

\noindent$\bullet$~\textbf{Unsupervised VOS.} This setting requires no manual input and aims to automatically segment primary objects in a video~\cite{fragkiadaki2015learning,yang2021dystab,RTNet,liu2021f2net,lu2020video,lu2020learning,Tokmakov2019,yang2019anchor,Li_2019_ICCV,wang2019learning2}, typically focusing on objects from pre-defined categories. Early methods relied on post-processing~\cite{fragkiadaki2015learning}, while later end-to-end methods became mainstream, broadly divided into local content encoding and contextual encoding. Local content encoding methods~\cite{jain2017fusionseg,cheng2017segflow,Li_2019_ICCV,li2018flow,Tokmakov2019,DBLP:conf/iccv/TokmakovAS17,zhou2020motion} often employ two-stream architectures to separately process optical flow and RGB information. Contextual content encoding methods~\cite{Lu_2019_CVPR,AGCNN,lu2021segmenting} aim to capture long-range dependencies and global context. Recent methods have adapted propagation frameworks for this task, DEVA~\cite{deva} proposes a decoupled framework combining image-level segmentation with class-agnostic temporal propagation, eliminating the need for task-specific video training data. EntitySAM~\cite{entitysam} extends SAM2 for zero-shot video entity segmentation by automatically discovering and tracking all entities without explicit prompts.

\subsection{Related Video Segmentation and Tracking Tasks} 
There are other video segmentation and tracking tasks related to VOS, \eg, video instance segmentation, video semantic segmentation, video panoptic segmentation, and video object tracking.

\noindent$\bullet$~\textbf{Video Instance Segmentation (VIS).} 
Video instance segmentation is extended from image instance segmentation by Yang \etal~\cite{yang2019video}, it simultaneously conducts detection, segmentation, and tracking of instances of predefined categories in videos. Thanks to the large-scale VIS dataset YouTube-VIS~\cite{yang2019video}, a series of learning methods have been developed and greatly advanced the performance of VIS~\cite{TPRPAMI,VMT,MaskFreeVIS,ying2023ctvis,dvis,dvis_daq,dvis++}. Then, occluded video instance segmentation is proposed by~\cite{OVIS} to study the VIS under occluded scenes. Similar to~\cite{OVIS}, we study video segmentation under complex scenarios like occlusions, but different from~\cite{OVIS}, we focus on video object segmentation, and the proposed \ournewdataset{2} dataset contains more videos and covers a broader range of real-world challenges beyond occlusion.

\noindent$\bullet$~\textbf{Video Semantic Segmentation (VSS).} 
Driven by the success in image semantic segmentation~\cite{long2015fully,Deeplabv2,ding2018context} and large-scale video semantic segmentation datasets~\cite{brostow2009semantic,cordts2016cityscapes,miao2021vspw}, video semantic segmentation has drawn lots of attention and achieved significant achievements. Compared to image domain, temporal consistency and model efficiency are the new efforts in the video domain. For example, Sun \etal~\cite{sun2022coarse,sun2024learning} propose Coarse-to-Fine Feature Mining to capture both static context and motional context.
Syed Hesham \etal~\cite{TV3S} propose a state space model-based~\cite{mamba} architecture for efficient temporal feature sharing.

\noindent$\bullet$~\textbf{Video Panoptic Segmentation (VPS).} Kim \etal~\cite{kim2020video} introduce panoptic segmentation to the video domain to simultaneously segment and track both the foreground instance objects and background stuff. They also build Cityscapes-VPS dataset with 500 videos. Then, Miao \etal~\cite{VIPSeg} build a larger VPS dataset called VIPSeg with 3,536 videos. Existing methods~\cite{Woo_2021_CVPR,Qiao_2021_CVPR} mainly add temporal refinement or cross-frame association modules upon image panoptic segmentation models~\cite{xiong2019upsnet} to enhance temporal conformity and instance tracking performance. Li \etal~\cite{OMGSeg} propose OMG-Seg, a unified transformer-based model that supports video panoptic segmentation along with over ten other segmentation tasks via task-specific queries and outputs.

\noindent$\bullet$~\textbf{Video Object Tracking (VOT).} 
Different from VOS that focuses on segmentation, VOT~\cite{vot_survey} aims to locate a target object with bounding boxes in subsequent frames given its initial bounding box annotation.
VOT has seen significant progress in recent years, with methods designed to handle challenging scenarios such as scale variations, occlusions, distractors, and complex backgrounds.
The dominant approaches can be broadly categorized into Siamese-based methods~\cite{siamese,zhang2019deeper,guo2020siamcar} that learn discriminative feature embeddings through twin networks, and transformer-based methods~\cite{cui2022mixformer,stark,tt_vot,lin2022swintrack,gat_tracking} that leverage self-attention mechanisms to model long-range dependencies for robust tracking.
These methods have achieved impressive performance on existing VOT benchmarks like VOT~\cite{vot2022}, LaSOT~\cite{lasot}, and GOT-10k~\cite{got-10k}. 
The proposed \ournewdataset{2} dataset also supports VOT task while introducing more complex real-world scenarios like dense crowds, occlusions, and frequent disappearance-reappearance that pose significant challenges to existing tracking methods.

\subsection{Complex Scene Understanding}
Complex scene understanding has become a research focus in the image understanding domain~\cite{ding2020semantic,lazarow2020learning,deocclusion,compositional2,PMOSR,repulsion,orcnn,chiou2021recovering,li2023transformer,wu2023open,miao2022region,zhan2022tri}. For example, Ke \etal~\cite{bcnet} propose Bilayer Convolutional Network (BCNet) to decouple overlapping objects into occluder and occludee layers. Zhang \etal~\cite{deocclusion} propose a self-supervised approach to conduct de-occlusion by ordering recovery, amodal completion, and content completion. 
On the video domain, however, occlusion understanding is still underexplored with only several multi-object tracking works \cite{mot_attn_occ1,mot_attn_occ2,mot_topo_occ1,mot_topo_occ2}. For example, Chu \etal~\cite{mot_attn_occ1} propose a spatial temporal attention mechanism (STAM) to capture the visible parts of targets and deal with the drift brought by occlusion. Zhu \etal~\cite{mot_attn_occ2} propose dual matching attention networks (DMAN) to deal with the noisy occlusions in multi-object tracking. Li \etal~\cite{TETr} propose to track every thing in the open world by performing class-agnostic association. In this work, we build a new complex video object segmentation dataset, \ournewdataset{2}, to facilitate future research on complex scene understanding in VOS and other related video understanding tasks.

\section{\ournewdataset{2} Dataset}\label{sec:OVOS_Dataset}

In this section, we introduce the newly built \ournewdataset{2} dataset. We first present the video collection and annotation process in \Cref{sec:videoannotation}, followed by dataset statistics and analysis in \Cref{sec:datasetStatistics}. Finally, we report the refined evaluation metrics in \Cref{sec:Metrics}.

\subsection{Video Collection and Annotation}\label{sec:videoannotation}

\myparagraph{Video Collection.} 
The videos in \ournewdataset{2} are obtained from two sources. The first source is inherited from \ournewdataset{1}~\cite{MOSEv1} with 2,149 videos. 
The second source consists of 2,875 newly self-captured videos from real-world scenarios and copyright-free videos from the internet that have not appeared in any existing dataset. 
MOSE is specifically designed for video object segmentation in complex scenes. To ensure the complexity and diversity of the collected videos, we follow a set of strict selection rules:
\begin{itemize}
\setlength\itemsep{0.16em}
    \item[R1.] Each video should contain multiple objects, except for challenging cases (\eg, camouflage). Specifically, videos with crowded objects of similar appearance are highly valued.
    \item[R2.] Occlusions are encouraged. Videos with occlusions, particularly those caused by other moving objects, are preferred.
    \item[R3.] Great emphasis should be placed on scenarios where objects disappear and then reappear due to occlusions or out-of-view.
    \item[R4.] The target objects should encompass a diverse range of scales (\eg, small-scale, large-scale) and visibility conditions (\eg, conspicuous, partially visible).
    \item[R5.] The video must exhibit clear motion, either from object movement or camera motion. Videos with static objects and a stationary camera should be discarded.
\end{itemize}
Besides the points mentioned above, we further emphasize the following rules in the design of \ournewdataset{2}:
\begin{itemize}
\setlength\itemsep{0.16em}
    \item[R6.] Target object categories should be diversified, including novel classes not present in \ournewdataset{1}, such as camouflaged objects, shadows, and reflections.
    \item[R7.] Longer videos are preferred for containing more challenging patterns, such as long-term occlusions, complex motion dynamics, and repeated object disappearance-reappearance, rather than merely for their duration.
    \item[R8.] A wide range of challenging environments is prioritized during collection, such as low-light scenes, cluttered scenes, and varying weather conditions (\eg, rain, fog, snow).
    \item[R9.] Multi-shot videos are encouraged, where objects undergo significant spatial or appearance changes across shots.
    \item[R10.] Videos requiring specific knowledge are deliberately included, such as optical character recognition, spatial reasoning, physical principles, and multi-view understanding.

\end{itemize}
\myparagraph{Video Annotation.} After collecting videos for \ournewdataset{2}, our research team manually reviews them to select suitable targets-of-interest for each video. We slightly trim the beginning and end of videos to reduce low-motion or simple frames. Next we annotate the first-frame masks of the selected targets. Then, the videos along with their first-frame masks are sent to the annotation team for annotation of the subsequent video frames.
    
\begin{table*}[t]
\footnotesize
\begin{center}
\caption{Statistical comparison between \ournewdataset{2} and existing video object segmentation and tracking datasets. ``Annotations'' denotes the number of annotated masks or boxes. ``Duration'' denotes the total duration of annotated videos, in minutes by default unless noted. 
``Disapp. Rate'' measures the frequency of objects disappearing in at least one frame, while ``Reapp. Rate'' measures the frequency of objects that previously disappeared and later reappear.
``Distractors'' quantifies scene crowding as the average number of visually similar objects per target in the first frame. * Unless otherwise specified, SA-V uses the combination of manual and auto annotations.
}
\label{tab:dataset-cmp}
\vspace{-6pt}
\renewcommand\arraystretch{1.2}
\setlength{\tabcolsep}{5.56pt}
\begin{tabular}{r|c|c|c|r|c|c|c|c|c|c}
\thickhline
\rowcolor{mygray}Dataset~~~~& Year&Videos & Categories& Objects &Annotations & Duration& Frames & Disapp. Rate & Reapp. Rate & Distractors\\
\hline
\hline
\multicolumn{11}{c}{\textbf{\textit{Video Object Tracking (VOT) Dataset}}} \\
\hline
{GOT-10k}~\cite{got-10k}    &2019   &\ \ 9,695& 563  &   10,200   &  \ \ \ \ 1.5M      &  \ \ \ \ 40.0 hr       & \ \ \ \ \ 1.5M &    \ \ 2.1\%   & \ \ 2.1\%  & \ \ 3.1\\
\rowcolor{mygray2}{LaSOT}~\cite{lasot}        &2019   &\ \  1,500&\ \  85    &   2,148   & \ \ \ \ \ 3.9M        & \ \ \ \ 35.8 hr     & \ \ \ \ \  3.9M  & 17.1\%  & 16.9\%  & \ \ 3.4\\
{VOT}~\cite{vot2022}    &2022   &\ \ \ \ \ \ 62   &\ \  -   &\ \ \ \  62    & \ \ 19,826        & \ \ \ \ \ \ 11.10   & \ \ \ 19,903  &  19.4\%   &  17.7\%  &  \ \ 5.2\\
\rowcolor{mygray2}{DiDi}~\cite{dam4sam}        &2025   &\ \ \ \   180  &\ \  -   &\ \ 180 &  268,084  &  \ \ \ \ 152.71   & \ 274,882 & 40.0\%       & 40.0\%  & 10.6 \\
\hline
\multicolumn{11}{c}{\textbf{\textit{Video Object Segmentation (VOS) Dataset}}} \\
\hline
{SegTrack-v2}~\cite{li2013video}  &2013   &\ \ \ \ \ \ 14     &\ \ 11 &\ \ \ \ 24     & \ \ \ \  1,475     & \ \ \ \ \ \ \  0.59 & \ \ \ \ \ \ \ 947 &\ \ 8.3\%   &\ \  0.0\%  & \ \ 5.4\\ 
\rowcolor{mygray2}{{YouTube-Objects}}~\cite{jain2014supervoxel} &2014   &\ \ \ \ 126     &\ \ 10 &\ \ \ \ 124     & \ \ \ \ 2,092      & \ \ \ \ \ \ \  9.01 &  \ \ \ \ 2,127 & \ \ 6.5\%   &  \ \ 1.6\% & -  \\  
{FBMS}~\cite{DBLP:journals/pami/OchsMB14}           &2014   &\ \ \ \ \ \ 59     &\ \ 16 &\ \ 139    &  \ \ \ \ 1,465     &  \ \ \ \ \ \ \ 7.70 & \ \ 13,860 & 11.2\%  & - & - \\
\rowcolor{mygray2}{JumpCut}~\cite{JumpCut}            &2015   &\ \ \ \ \ \ 22     &\ \ 14 &\ \ \ \ 22     & \ \ \ \  6,331     & \ \ \ \ \ \ \ 3.52 & \ \ \ \ 5,315 &\ \ 0.0\%   &\ \  0.0\%  & - \\
{DAVIS}$_{16}$\cite{davis2016}                      &2016   &\ \ \ \ \ \ 50     &\ \ -  &\ \ \ \ 50     & \ \ \ \  3,455     & \ \ \ \ \ \ \  2.88 & \ \ \ \ 3,440 &  11.1\%  & \ \ 4.9\% & \ \ 2.6\\
\rowcolor{mygray2}{DAVIS}$_{17}$\cite{davis2017}    &2017   &\ \ \ \ \ \ 90     &\ \ -  &\ \ 205    & \ \  13,543    & \ \ \ \ \ \ \ 5.17& \ \ \ \ 6,208 & 16.1\%    & 10.7\% & \ \ 3.7\\
{YouTube-VOS}~\cite{youtube_vos}                    &2018   &\ \ 4,453  &\ \ 94 &7,755  & 197,272   & \ \ \ 334.81& 120,532 & 13.0\%    &\ \  8.0\% & \ \ 3.0\\
\rowcolor{mygray2}{VOTS}~\cite{vots}                &2023   &\ \ \ \ 144     &\ \   - &\ \ 341      &  -     & \ \ \ 166.00  & 298,640 &  - &  -  &  -   \\
{VOST}~\cite{vost}                &2023   &\ \ \ \ 713    & 154  & 1,726      &    173,758     &  \ \ \ 251.92 & \ \ 75,547 & 46.5\%  &  44.4\%  & \ \ 5.3\\
\rowcolor{mygray2}{LVOSv1}~\cite{LVOSv1}            &2023   &\ \ \ \ 220    &\ \ 27 &\ \  282     &  156,432         & \ \ \ 351.00  & 126,280 &  50.0\% & 46.7\% & \ \ 3.7\\
{LVOSv2}~\cite{LVOSv2}                              &2024   &\ \ \ \ 720      &\ \  44  &  1,132     & 407,945           & \ \ \ 823.00   & 296,401 &  36.1\% & 32.5\% & \ \ 4.6\\
\rowcolor{mygray2}{SA-V*}~\cite{SAM2}                &2024   &50,900  &\ \ -  & 642,600 & \ \ 35.5M   & \ 196.0 hr& \ \ \ 4.2M & 58.7\%   & 27.7\% & \ \ 6.2\\
\hline
{{\ournewdataset{1}}}~\cite{MOSEv1} &{2023}&\ \ {2,149} &\ \  {36} & {5,200} & {431,725} & \ \ \ {443.62} & 130,149 &{41.5\%} & 23.9\% & \ \ 6.5\\
\rowcolor{mygray}\textbf{{\ournewdataset{2}}\!~\ \ \ \ }  &\textbf{2025}&\ \ 
\textbf{\numvideo} & \textbf{\numclass} & \textbf{\numobject} & \textbf{\nummask} & \textbf{\numduration} & \textbf{\textbf{\numframes}}  &\textbf{61.8\%}  & \textbf{50.3\%} &  \textbf{13.6} \\
\hline
\end{tabular}
\end{center}
\vspace{-3.6mm}
\end{table*}

Using the given first-frame mask as a reference, annotators are required to identify the corresponding target and then track and annotate its segmentation masks across all subsequent frames. To facilitate this process, an interactive annotation tool is developed to automatically load videos and target objects. Annotators can preview the video and first-frame mask, annotate and visualize masks in later frames, and save the results. The annotation tool also has a built-in interactive video object segmentation model SAM2~\cite{SAM2} to assist annotations in producing high-quality masks.
To ensure annotation quality in complex scenes, annotators are required to consistently track the target and provide precise segmentation. For frames where the target disappears or is fully occluded, the masks must remain empty. All videos are annotated at a minimum of 5 FPS, while a subset is annotated at full FPS to evaluate the frame-rate robustness of VOS models.

After annotation, all videos are carefully reviewed by our verification team to ensure high-quality masks.

\subsection{Dataset Statistics}\label{sec:datasetStatistics}
In~\Cref{tab:dataset-cmp}, we analyze the data statistics of \ournewdataset{2} in comparison with existing VOS datasets, such as DAVIS~\cite{davis2016,davis2017}, YouTube-VOS~\cite{youtube_vos}, LVOS~\cite{LVOSv1,LVOSv2}, SA-V~\cite{SAM2}, \ournewdataset{1}\cite{MOSEv1}, as well as VOT datasets, including GOT-10k\cite{got-10k}, LaSOT~\cite{lasot}, VOT~\cite{vot2022}, and DiDi~\cite{dam4sam}.
\ournewdataset{2} expands \ournewdataset{1} by adding 2,875 new videos, reaching a total of \numvideo videos and \nummask mask annotations for \numobject objects.

\myparagraph{Categories.}
\ournewdataset{2} contains \numclass object categories, the largest among existing VOS datasets. \cref{fig:categories} presents the detailed category distribution of MOSE. Building on the 36 categories of \ournewdataset{1}, \ournewdataset{2} significantly expands the scope to \numclass, covering not only common categories such as squirrels, footballs, and otters, but also rare ones like Newton’s cradle and camouflaged objects, as well as non-physical targets like shadows. This extensive coverage enables more comprehensive and robust evaluation of VOS methods.

\myparagraph{Disappearance-Reappearance.} \ournewdataset{2} significantly surpasses its predecessor \ournewdataset{1} in terms of object disappearance and reappearance. The ``Disapp.~Rate'' increases from 41.5\% to 61.8\%, while the ``Reapp.~Rate'' more than doubles from 23.9\% to 50.3\%. \ournewdataset{2} also exceeds SA-V (58.7\%) among VOS datasets and DiDi (40.0\%) among VOT datasets in ``Disapp. Rate'', while its 50.3\% ``Reapp.~Rate'' outperforms LVOSv1 (46.7\%) and DiDi (40.0\%). These characteristics make \ournewdataset{2} the most challenging dataset for studying disappearance-reappearance scenarios.

\begin{table}[t]
    \centering
    \caption{Occlusion rate comparison among different datasets.}
    \vspace{-2mm}
    \label{tab:occlusion}
    \renewcommand\arraystretch{1.06}
    \setlength{\tabcolsep}{8.pt}
    \begin{tabular}{r|c|c|c|c}
    \thickhline
         \rowcolor{mygray}{Dataset}  & Mean & mBOR & mAOR &  mMLLMOR  \\
         \hline
         \hline
         DAVIS$_{17}$~\cite{davis2017}                    & 20.6  & \ \ 3.4     & 23.7            & 34.6  \\
         \rowcolor{mygray2}YouTube-VOS~\cite{youtube_vos} & 23.2  & \ \ 5.7     & 26.0            & 38.0  \\
         LVOSv2~\cite{LVOSv2}                             & 25.4  & \ \ 8.4     & 30.6            & 37.2      \\
         \rowcolor{mygray2}SA-V~\cite{SAM2}               & 36.1  & 27.4        & 37.2            & 43.6      \\
         \hline
         MOSEv1~\cite{MOSEv1}                             & 36.4  & 23.7        & 41.2            & 44.2      \\
         \rowcolor{mygray}\textbf{MOSEv2\!~\ \ \ \ }          & \textbf{47.0}       & \textbf{28.3}   & \textbf{54.8}            & \textbf{57.8}  \\
         \hline
    \end{tabular}
    \vspace{-1.6mm}
\end{table}

\begin{figure}[t]
    \centering
    \includegraphics[width=0.98\linewidth]{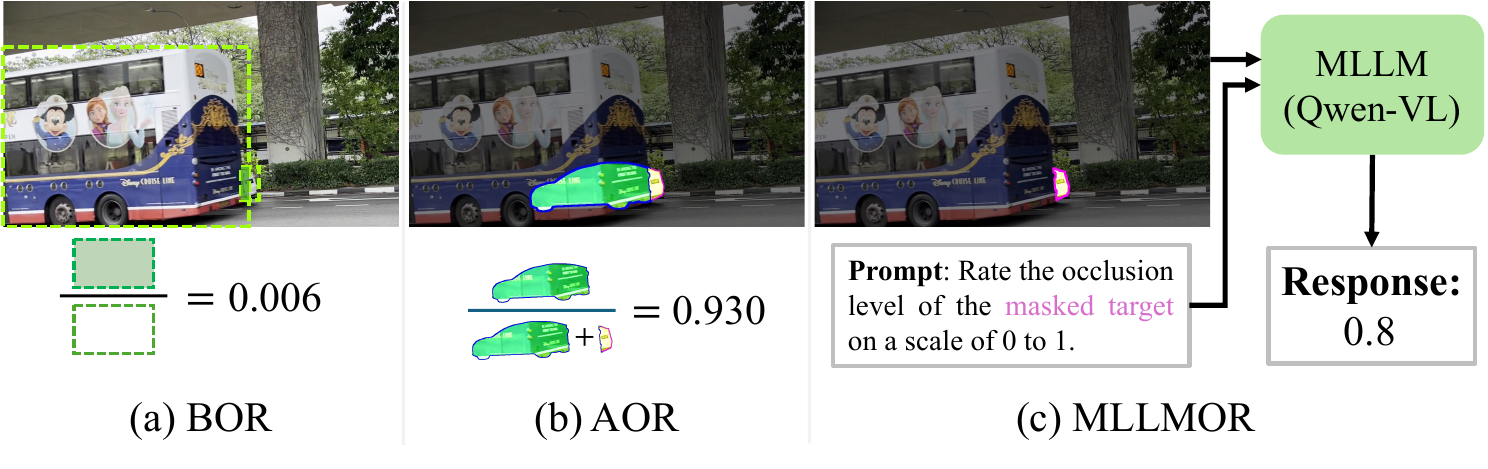}
    \vspace{-2mm}
    \caption{Occlusion evaluation protocol. (a) BOR: Bounding-box Occlusion Rate~\cite{OVIS}, (b) AOR: Amodal-mask Occlusion Rate, (c) MLLMOR: MLLM-assisted Occlusion Rate.}
    \label{fig:occlusion}
    \vspace{-2mm}
\end{figure}

\myparagraph{Crowding.}
To assess crowding complexity, we compute the ``Distractors'' metric, which quantifies the average number of visually similar objects per target in the first frame, using T-Rex2~\cite{trex2}. 
\ournewdataset{2} reaches 13.6 distractors per object, more than twice that of \ournewdataset{1} (6.5), and also higher than SA-V (6.2) and LVOSv2 (4.6). Remarkably, it even surpasses DiDi~\cite{dam4sam} (10.6), a dataset specifically designed to emphasize distractors in VOT, underscoring MOSEv2's complexity and its significance for advancing robust perception in densely crowded scenes.

\myparagraph{Occlusion.} We compare occlusion levels of \ournewdataset{2} with other datasets in \Cref{tab:occlusion}. While \ournewdataset{2} achieves the highest mBOR score~\cite{OVIS} of 28.3, this metric offers only a coarse estimation~\cite{MOSEv1}. As shown in \cref{fig:occlusion} (a), an object may be heavily occluded yet still yield a BOR near zero. To address this limitation, we introduce two complementary metrics: Amodal-mask Occlusion Rate (AOR) and MLLM-assisted Occlusion Rate (MLLMOR), shown in \cref{fig:occlusion} (b) and (c), respectively. AOR measures the ratio between visible and amodal mask areas generated by the amodal segmentation model DiffVAS~\cite{chen2025diffvas}. MLLMOR leverages a multimodal large language model (we use Qwen2.5-VL-32B\cite{qwenvl25}) to assess occlusion severity. 
We compute the final occlusion estimate as the average of all three metrics. As shown in \Cref{tab:occlusion}, \ournewdataset{2} achieves a mean occlusion rate of 47.0, substantially exceeding \ournewdataset{1} (36.4) and SA-V (36.1), establishing \ournewdataset{2} as the most challenging dataset for studying occlusion in videos.

\begin{figure*}[t!]
    \centering
    \begin{minipage}[t]{0.25\textwidth}
        \centering
        \includegraphics[width=\linewidth]{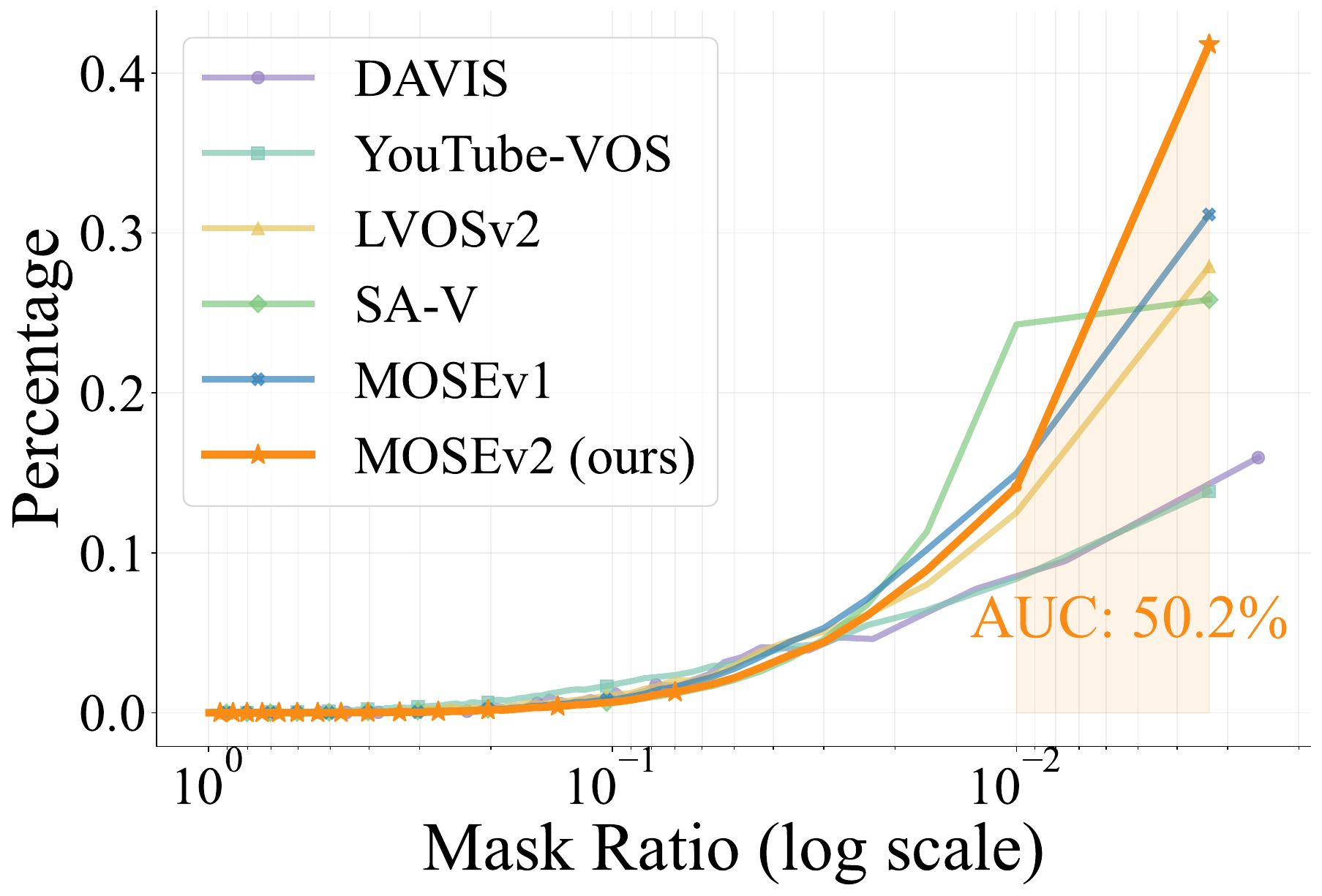}
        \vspace{-6mm}
        \caption{Mask size distribution, normalized by video resolution.}
        \label{fig:mask_ratio}
    \end{minipage}
    \hfill
    \begin{minipage}[t]{0.52\textwidth}
        \centering
        \includegraphics[width=\linewidth]{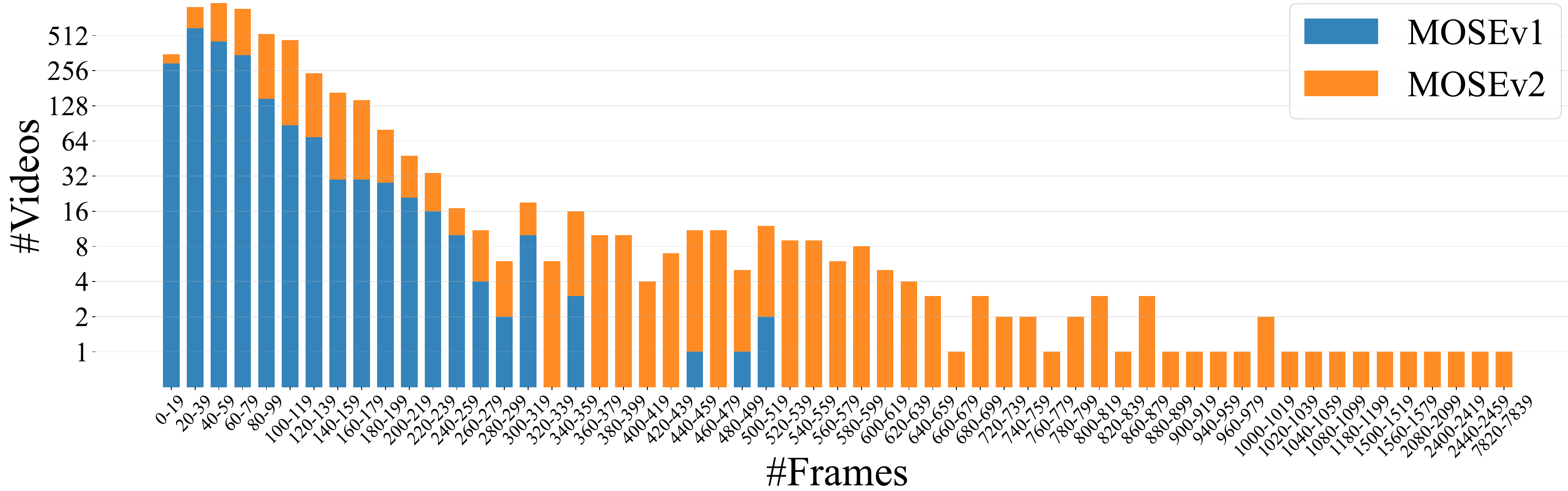}
        \vspace{-6mm}
        \caption{Video length distributions. Compared to \ournewdataset{1}, \ournewdataset{2} includes more long videos, with the longest reaching 7,825 frames.}
        \label{fig:length}
    \end{minipage}
    \hfill
    \begin{minipage}[t]{0.20\textwidth}
        \centering
        \includegraphics[width=\linewidth]{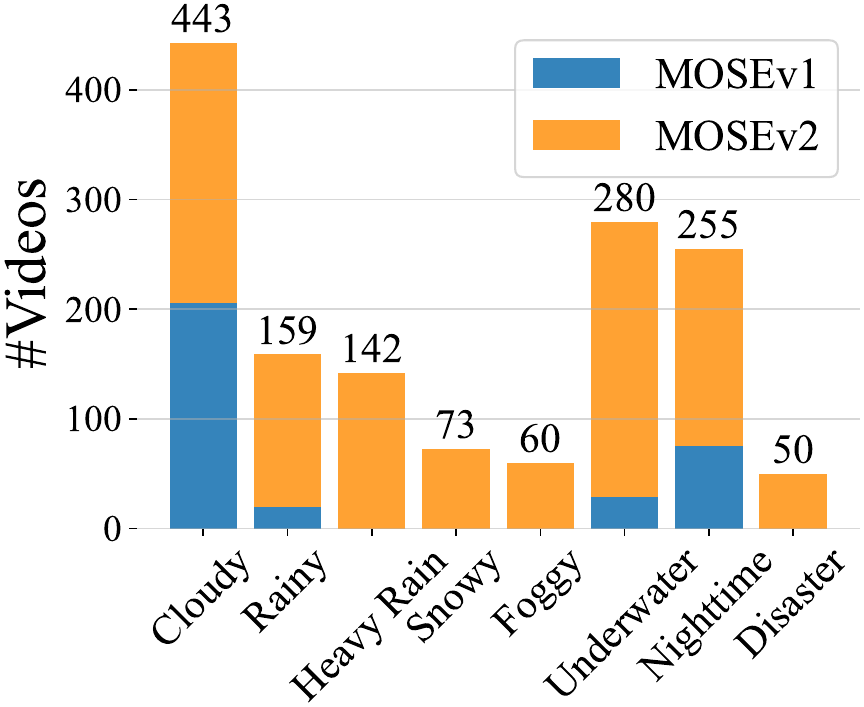}
        \vspace{-6mm}
        \caption{Challenging environment distribution.}
        \label{fig:environment}
    \end{minipage}
    \vspace{-2mm}
\end{figure*}

\noindent\textbf{Mask Size.}
\cref{fig:mask_ratio} compares the distribution of mask sizes (normalized by video resolution) across datasets. \ournewdataset{2} contains a substantially higher proportion of small masks (size < 0.01), reaching 50.2\%, significantly above DAVIS (25.3\%), YouTube-VOS (18.4\%), LVOSv2 (34.8\%), SA-V (40.7\%), and \ournewdataset{1} (39.5\%). This high prevalence of small objects poses greater challenges for fine-grained perception and accurate segmentation.

\myparagraph{Video Length.}
\cref{fig:length} presents the video length distribution in \ournewdataset{2}. Compared to only 11 videos exceeding 300 frames (around 1 minute) in \ournewdataset{1}, \ournewdataset{2} provides 183 such long videos, with the longest reaching 7,825 frames, around 26 minutes. The average video length increases from 60.6 to 93.2 frames, enabling more comprehensive evaluations of long-term temporal consistency and tracking robustness. While LVOSv2~\cite{LVOSv2} includes 362 videos over 300 frames with an average length of 590.9 frames, our 183 long videos average 598.4 frames and extend up to 7,825 frames, far beyond LVOSv2’s maximum of 2,280. Importantly, video length alone does not imply difficulty. In \ournewdataset{2}, long videos are not included merely for their duration, but intentionally designed to include richer dynamics and more complex scenarios, such as object disappearance, occlusion, scene transitions, and multi-shot clips. For example, LVOSv2's reappearance rate is only 32.5\%, substantially lower than our 50.3\%, highlighting the increased complexity in \ournewdataset{2}.

\myparagraph{Complex Environments.}
\cref{fig:environment} shows the distribution of challenging environmental conditions. Compared to \ournewdataset{1}, \ournewdataset{2} significantly expands the coverage of adverse scenarios. For example, rainy videos increased from 20 to 159, and underwater scenes from 29 to 280. \ournewdataset{2} also introduces new conditions not present in \ournewdataset{1}, including 142 heavy rain, 73 snow, 60 fog, and 50 disaster scenarios (\eg, earthquake, flood). In total, \ournewdataset{2} provides 443 cloudy, 159 rainy, 142 heavy rain, 73 snowy, 60 foggy, 280 underwater, 255 nighttime (\vs 75 in \ournewdataset{1}), and 50 disaster videos. This substantial expansion establishes \ournewdataset{2} as a more comprehensive dataset for exploring model robustness under diverse complex environments.

\begin{figure}[t!]
    \centering
    \includegraphics[height=2cm]{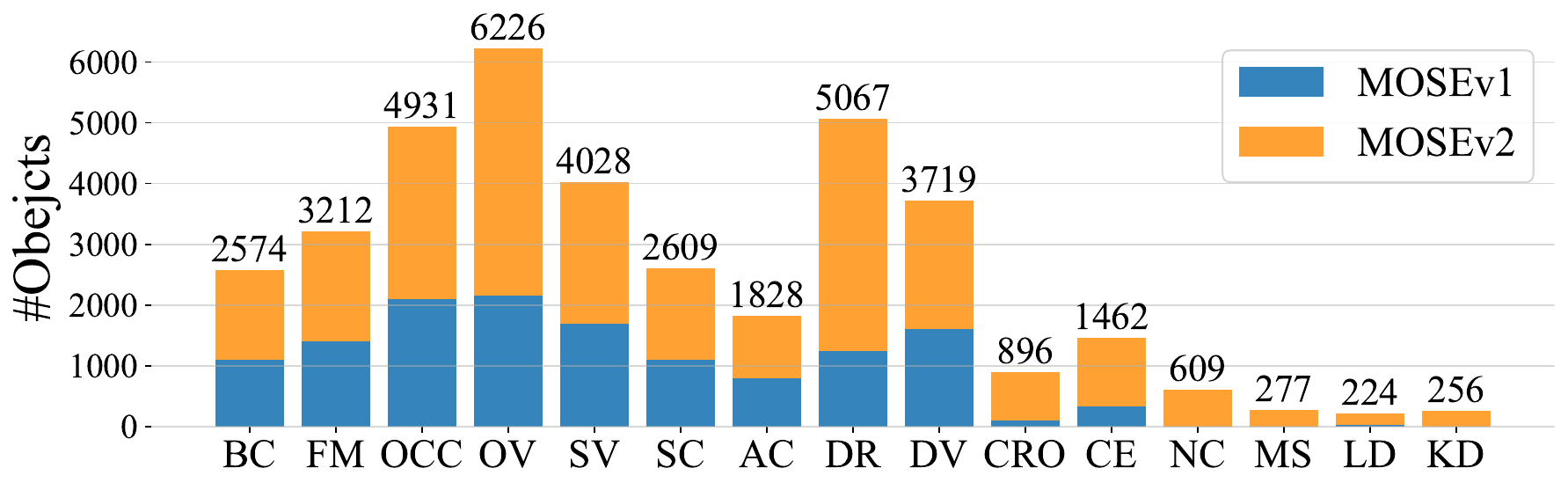}
    \includegraphics[height=2cm]{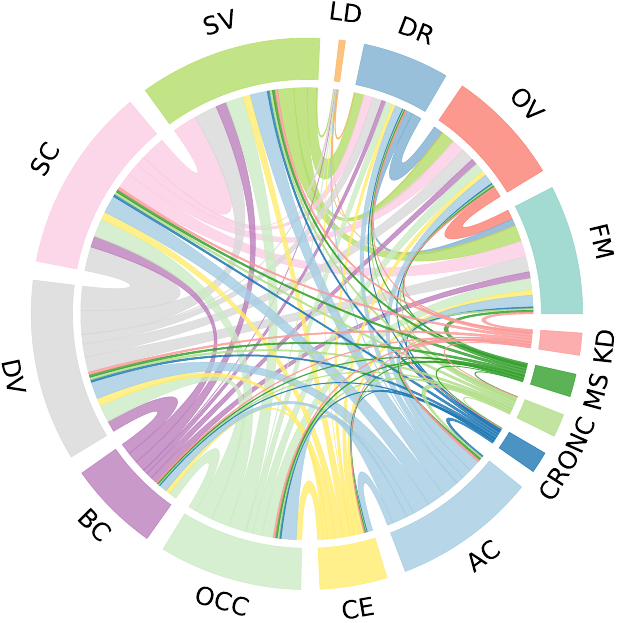}
    \vspace{-2mm}
    \caption{(Left) Distribution of objects attributes. (Right) Attribute correlations in \ournewdataset{2}.}
    \label{fig:attribute}
    \vspace{-2mm}
\end{figure}

\myparagraph{Object Attribute Analysis.}
Following DAVIS~\cite{davis2017}, we define 15 object attributes for \ournewdataset{2} in \Cref{tab:challenge}. As shown in \cref{fig:attribute} (left), \ournewdataset{2} greatly expands the coverage of challenging scenarios compared to \ournewdataset{1}. For example, objects with occlusion (OCC) increase from 2,100 to 4,931, disappearance-reappearance (DR) from 1,243 to 5,067, complex environments (CE) from 330 to 1,462, and long duration (LD) from 33 to 224. In addition, \ournewdataset{2} introduces new attributes such as novel categories (NC, 609 objects), multi-shot sequences (MS, 277), and knowledge dependency (KD, 256). As shown in \cref{fig:attribute} (right), a chord diagram illustrates the co-occurrence patterns between attributes, offering insights into the interplay of real-world challenges. This comprehensive attribute set makes \ournewdataset{2} a more rigorous benchmark for evaluating model under diverse and complex conditions.

\begin{table}[t]
	\centering
    \caption{Definitions of object attributes in \ournewdataset{2}. We adopt part of the attributes from DAVIS~\cite{davis2017} (top) and extend them with additional complex video attributes (bottom).}
	\vspace{-2mm}
    \label{tab:challenge}
    \renewcommand\arraystretch{1.2}
    \setlength{\tabcolsep}{3.06pt}
	\begin{tabular}{l|l}
        \thickhline
        \rowcolor{mygray}Attr. & Definition  \\
        \hline
        \hline
        BC        &  \textit{Background Clutter.} The background and the target object exhibit \\  
        & similar visual appearances.  \\
        \rowcolor{mygray2}FM        & \textit{Fast Motion.} The average, per-frame object motion, computed as \\ 
        \rowcolor{mygray2}&  centroids Euclidean distance, is larger than $\tau_{fm} = 20$ pixels.\\
        OCC       & \textit{Occlusion.} The target object is partially or fully occluded in video. \\
        \rowcolor{mygray2}OV        & \textit{Out-of-view.} The target object leaves the video frame completely.\\
        SV        & \textit{Scale Variation.} The ratio of any pair of bounding boxes is outside  \\
        & of range {[}0.5,2.0{]}.                  \\
        \rowcolor{mygray2}SC        & \textit{Shape Complexity.} The target exhibits complex boundary structures.  \\
        AC        & \textit{Appearance Change.} Significant appearance change, due to rotations   \\ 
        & and illumination changes. \\
        \hline
        \rowcolor{mygray}DR       & \textit{Disappearance-Reappearance.} The target object reappears after dis- \\
        \rowcolor{mygray}&  appearing in the video. \\
        DV        & \textit{Diverse Visibility.} The target object is small, inconspicuous, or \\
        & camouflaged in the scene. \\
        \rowcolor{mygray}CRO      & \textit{Crowding.} Multiple similar objects appear in close proximity. \\
        CE       & \textit{Complex Environment.} Object under challenging conditions such \\
        & as underwater, nighttime, and adverse weather (rain, snow). \\
        \rowcolor{mygray}NC       & \textit{Novel Categories.} Novel object categories, especially camouflaged \\
        \rowcolor{mygray}& objects and non-physical objects. \\
        MS       & \textit{Multi-Shots.} The object sequence contains multiple camera shots. \\
        \rowcolor{mygray}LD       & \textit{Long Duration.} Object duration exceeds 1 minute (300 frames). \\
        KD       & \textit{Knowledge Dependency.} Objects requiring specific knowledge \\
        &(e.g., OCR, spatial reasoning) for precise segmentation in videos. \\
        \thickhline          
		\end{tabular} 
        \vspace{-2mm}
\end{table}

\subsection{Evaluation Metrics}\label{sec:Metrics}
Following previous VOS works~\cite{davis2016,davis2017,MOSEv1}, we compute the region similarity \J and the contour accuracy \F as evaluation metrics. Given a predicted mask ${\hat{M}}\!\in\!\{0,1\}^{H\times W}$ and a ground-truth mask ${M}\!\in\!\{0,1\}^{H\times W}$, region similarity \(\mathcal{J}\) is computed as the Intersection-over-Union of ${\hat{M}}$ and ${{M}}$, \ie, $\mathcal{J} = ({{\hat{M} \cap  M}})/({{\hat{M}\cup M}})$.
To measure the contour quality of ${\hat{M}}$, contour recall $\text{R}_c$ and precision $\text{P}_c$ are calculated via bipartite graph matching~\cite{martin2004learning}. Then, the contour accuracy \(\mathcal{F}\) is the harmonic mean of the contour recall $\text{R}_c$ and precision $\text{P}_c$, \ie, $\mathcal{F} = {{2 \text{P}_c \text{R}_c}}/({{\text{P}_c + \text{R}_c}})$, which represents how closely the contours of predicted masks resemble the contours of ground-truth masks. Next, the average region similarity \(\mathcal{J}_{mean}\) and contour accuracy \(\mathcal{F}_{mean}\) over all objects are calculated as the final results. For brevity, we use \(\mathcal{J}\) and \(\mathcal{F}\) to represent \(\mathcal{J}_{mean}\) and \(\mathcal{F}_{mean}\), respectively. The overall performance is measured by $\mathcal{J}\&\mathcal{F}=(\mathcal{J}+\mathcal{F})/2$.

\begin{figure}[t]
    \centering
    \includegraphics[width=0.98\linewidth]{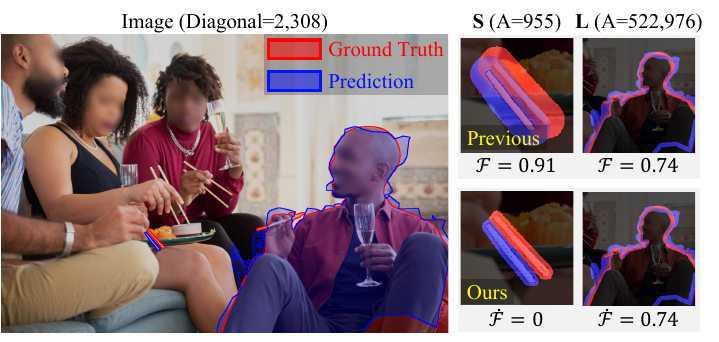}
    \vspace{-2mm}
    \caption{Comparison of the commonly used \F and our adaptive \Fn on \textbf{S}mall and \textbf{L}arge objects. For small objects, such as a chopstick with 955 pixels (only 0.039\% of the image area), \F yields exaggerated scores due to fixed resolution-based thresholds, while \Fn offers a more reliable measure by accounting for object scale. For large objects, such as a person (522k pixels, 21\% of the image area), \Fn remains consistent with \F, both yielding 0.74.}
    \label{fig:boundary}
\end{figure}

\myparagraph{Revisiting the $\mathcal{F}$ Score.}
The widely used $\mathcal{F}$ score has clear limitations for small objects, which is particularly problematic for \ournewdataset{2} that contains a large number of small targets.
Previous works~\cite{davis2016,davis2017,MOSEv1} adopt a fixed boundary threshold $w = 0.008 \times D$ for images of the same resolution, where $D$ is the image diagonal. While effective for images with uniform object sizes, this threshold ignores object scale and biases the evaluation of small objects. 
As shown in \cref{fig:boundary}, for a chopstick with only 955 pixels, merely 0.039\% of the image area, the predicted and ground-truth masks do not overlap, yet the fixed threshold excessively dilates boundaries and yields an inflated $\mathcal{F}$ score of 0.91.
To address this issue, we propose an adaptive boundary threshold:
\begin{equation}
\dot{w} = \min(0.008 \times D, \alpha \times \sqrt{A}),
\end{equation}
where $A$ is the object's area in pixels and $\alpha$ is a scaling factor. 
Based on boundary statistics from DAVIS and MOSE, we set $\alpha = 0.1$ to maintain reasonable boundary widths for average-sized objects while better handling small ones.
We denote this adaptive-threshold-based metric as \Fn, which provides a fairer boundary evaluation across different object scales. For the small chopstick in \cref{fig:boundary}, the improved \Fn correctly assigns a score of 0, while for large object, such as a person with 522k pixels (21\% of the image area), \Fn maintains consistency with the original metric \F.

\begin{figure}[t]
    \centering
    \includegraphics[width=\linewidth]{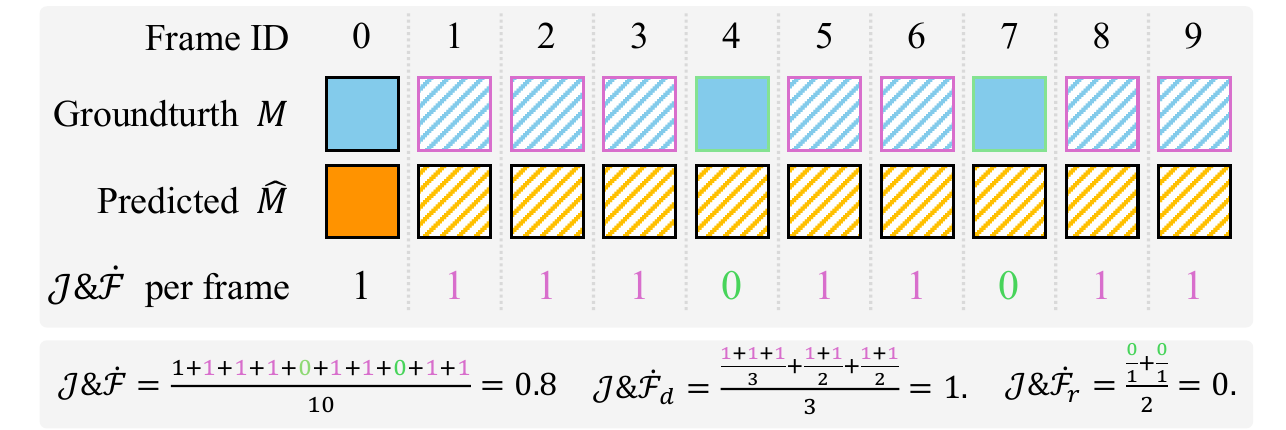}
    \vspace{-6.6mm}
    \caption{Illustration of \JFn, \JFnd, and \JFnr. The slashed boxes denote frames with empty masks. For simplicity, all predicted masks are assumed to match the ground truth perfectly.}
    \label{fig:jf}
    \vspace{-1mm}
\end{figure}

\myparagraph{Disappearance and Reappearance Metrics.}
Given the frequent object disappearance-reappearance in \ournewdataset{2}, we compute dedicated scores: $\mathcal{J}\&\mathcal{\dot{F}}_d$ for disappearance clips where the target is absent, and $\mathcal{J}\&\mathcal{\dot{F}}_r$ for reappearance clips where the target returns. 
As shown in \cref{fig:jf}, we first compute these metrics per disappearance or reappearance clip, average them to obtain sequence-level scores.
These metrics address a key limitation of the widely used $\mathcal{J}\&\mathcal{F}$, which are averaged over all frames in a video and thus biased by the proportion of empty-target frames. For example, in videos with many disappearance frames, models that tend to predict empty masks may appear strong, while in videos with few disappearance frames, models that blindly predict masks may benefit since errors on empty frames have little impact.

By isolating evaluation to disappearance and reappearance clips, $\mathcal{J}\&\mathcal{\dot{F}}_d$ and $\mathcal{J}\&\mathcal{\dot{F}}_r$ provide clearer insights: models that fail to suppress masks during disappearance perform poorly on $\mathcal{J}\&\mathcal{\dot{F}}_d$, while those unable to recover the target after its return are penalized on $\mathcal{J}\&\mathcal{\dot{F}}_r$. Only models handling both cases effectively achieve high scores on both metrics. For $\mathcal{J}\&\mathcal{\dot{F}}_r$, we deliberately exclude the initial continuous presence of the target, where reference is strongest, focusing instead on true reappearance after disappearance, which better reflects recovery under ambiguity.

\section{Experiments}

We conduct comprehensive experiments on the newly built \ournewdataset{2} dataset, benchmarking multiple video object segmentation settings, including semi-supervised VOS with mask, box, and point initialization, as well as unsupervised and interactive VOS. We further evaluate video object tracking methods on \ournewdataset{2}, demonstrating its broad applicability beyond segmentation.

\begin{table*}[t]
\centering
\renewcommand\arraystretch{1.12}
\setlength{\tabcolsep}{2.36pt}
\footnotesize
\caption{Benchmark results of mask-initialization semi-supervised VOS methods on \ournewdataset{2} validation set. ``ZS'' indicates that the model uses zero-shot evaluation. Inference speed (FPS) and GPU memory usage (GiB) are measured on a single A6000 GPU. For SAM2 and its variants, video frames are offloaded to CPU memory to balance inference speed and memory usage.}
\label{tab:svos}
\vspace{-2.16mm}
\begin{tabular}{l|c|cc|ccccccc|ccc|c|c|c|c}
\thickhline
\rowcolor{mygray3}    & &\multicolumn{2}{c|}{} &  \multicolumn{7}{c|}{\textbf{\ournewdataset{2}}}&  \multicolumn{3}{c|}{{\ournewdataset{1}}}&SA-V$_{test}$&LVOSv2&DAVIS$_{17}$ & YT-VOS$_{19}$ \\
\rowcolor{mygray3}\multirow{-2}{*}{Method}& \multirow{-2}{*}{Pub.}& \multirow{-2}{*}{FPS}& \multirow{-2}{*}{Mem.} & \JFn &\J & \Fn & \JFnd & \JFnr & \F & \JF &  \JF &\J  & \F &\JF & \JF & \JF & \JF\\ 
\hline
\hline
AOT-L~\cite{AOT} & \pub{NeurIPS'21} & 19.7 & 3.8 & 30.2 &29.0 & 31.4 & 67.8 & \ \ 7.8  & 32.9 & 31.0 &  57.2 &53.1 & 61.3 & 50.3 & 63.9 & 84.9 & 84.1 \\
STCN~\cite{STCN} & \pub{NeurIPS'21} & 45.1 & 6.2 & 29.7 &28.9 & 30.5 & 79.4 & \ \ 8.1  & 31.4 & 30.2 &  50.8 &46.6 & 55.0 & 62.5 & 60.6 & 85.4 & 82.7 \\
RDE~\cite{RDE}   & \pub{CVPR'24}    & 32.7 & 1.4 &  32.0 &30.7 & 33.3 & 62.7 & 12.6 & 35.0 & 32.8 &  48.8 &44.6 & 52.9 & 53.9 & 62.2 & 84.2 & 81.9 \\
XMem~\cite{xmem} & \pub{ECCV'22}    & 49.8 & 1.6  &  36.3 &34.7 & 37.9 & 56.6 & 14.8 & 40.0 & 37.4 &  57.6 &53.3 & 62.0 & 62.3 & 64.5 & 86.2 & 85.6 \\
DeAOT-L~\cite{DeAOT} & \pub{NeurIPS'22} & 21.2 & 3.7 &  32.6 &30.7 & 34.5 & 33.5 & 18.3 & 37.2 & 33.9 &  59.4 &55.1 & 63.8 & 61.8 & 63.9 & 85.2 & 86.0 \\
DEVA~\cite{deva}     & \pub{ICCV'23}    & 43.0 & 1.0     &  38.3 &36.6 & 40.0 & 55.1 & 18.5 & 42.2 & 39.4 &  60.0 &55.8 & 64.3 & 56.2 & -    & 87.0 & 85.4 \\
XMem++~\cite{xmem++} & \pub{ICCV'23}    & 30.1 & 1.4  &  34.2 &32.5 & 35.9 & 51.6 & 15.5 & 37.9 & 35.2 &  56.0 &51.5 & 60.6 & -    & -    & -    & -    \\
Cutie-B~\cite{cutie} & \pub{CVPR'24}    & 44.1 & 0.9     &  42.8 &41.1 & 44.4 & 64.5 & 18.3 & 46.8 & 43.9 &  69.9 &65.9 & 74.1 & 60.7 & -    & 87.9 & 87.0 \\
JointFormer~\cite{jointformer}& \pub{PAMI'25} &  \ \ 7.2  &  3.6     &  37.7 &36.0 & 39.4 & 57.3 & 18.3 & 41.1 & 38.6 &  70.2 &66.3 & 74.0 & -    & -    & 90.6 & 87.5 \\
\hline
SAM2-B+~(ZS)~\cite{SAM2} & \pub{ICLR'25} & 23.4 &  2.7     &  43.1 &41.4 & 44.8 & 60.6 & 20.6 & 47.0 & 44.2 &  73.6 &69.5 & 77.6 & 77.0 & 83.1 & 90.2 & 88.6 \\
SAM2-B+~\cite{SAM2}      & \pub{ICLR'25} & 23.4 &  2.7    &  46.0 &44.2 & 47.8 & 61.6 & 23.2 & 50.0 & 47.1 &  74.7 &70.6 & 78.8 & -    & -    & -    & -     \\
SAMURAI-B+~\cite{SAMURAI}       & \pub{Preprint'24} & 17.7 &  2.7     &  47.4 &45.3 & 49.5 & 45.9 & 33.6 & 52.2 & 48.8 &  73.3 &69.0 & 77.5 & -    & -    & -& -    \\
DAM4SAM-B+~\cite{dam4sam}  & \pub{CVPR'25}  &  17.3    & 2.7     &  47.9 &45.8 & 50.0 & 51.3 & 32.0 & 52.6 & 49.2 &  73.8 &69.5 & 78.0 & -    & -    & -    & -    \\
SAM2Long-B+~\cite{SAM2Long} & \pub{ICCV'25} &  \ \ 9.4    &  6.0    &  48.6 &46.7 & 50.5 & 58.4 & 29.2 & 52.8 & 49.7 &  74.7 &70.6 & 78.8 & 80.8 & 85.2 & -    & -    \\
\hline
SAM2-L~(ZS)~\cite{SAM2}       & \pub{ICLR'25}  & 14.4 & 3.6     &  49.4 &47.6 & 51.3 & 63.4 & 27.8 & 53.8 & 50.7 &  74.5 &70.5 & 78.4 & 78.4 & 84.0 & 90.7 & 89.3 \\
SAM2-L~\cite{SAM2}      & \pub{ICLR'25}        &  14.4 & 3.6    &  49.7 &47.9 & 51.5 & 64.5 & 27.1 & 53.8 & 50.9 &  76.4 &72.3 & 80.5 & -    & -    & -    & -    \\
SAMURAI-L~\cite{SAMURAI}        & \pub{Preprint'24} &  12.1 &  3.5    &  51.1 &49.0 & 53.2 & 52.4 & 34.9 & 55.8 & 52.4 &  75.6 &71.4 & 79.8 & -    & -    & -    & -    \\
DAM4SAM-L~\cite{dam4sam}        & \pub{CVPR'25}     &  12.3 &  3.5    &  51.2 &49.2 & 53.2 & 57.2 & 34.2 & 55.6 & 52.4 &  75.6 &71.5 & 79.8 & -    & -    & -    & -    \\
SAM2Long-L~\cite{SAM2Long}       & \pub{ICCV'25}    &  \ \ 7.1    & 6.8   &  51.5 &49.6 & 53.4 & 62.5 & 30.6 & 55.8 & 52.7 &  77.1 &73.0 & 81.2 & 81.2 & 85.3 & 88.8    & 90.2    \\
\thickhline
\end{tabular}
\vspace{-2.6mm}
\end{table*}

\myparagraph{Implementation Details.}
The proposed \ournewdataset{2} follows the same data format as MOSEv1~\cite{MOSEv1} and YouTube-VOS~\cite{youtube_vos}. 
For methods developed before SAM2, we ensure fair comparisons by replacing the YouTube-VOS~\cite{youtube_vos} training set with \ournewdataset{2} while strictly following the original YouTube-VOS training configurations.
These methods are trained with image-pretrained backbones without using any additional video datasets.
For SAM2-based models, we adopt SAM2.1 as the default initialization and fine-tune exclusively on \ournewdataset{2}.
We evaluate model performance using standard metrics (\J, \F, and \JF) on \ournewdataset{2} validation set, following the DAVIS protocol~\cite{davis2016,davis2017}.
To better capture the complex challenges in \ournewdataset{2}, we additionally report \Fn, \JFn, \JFnd, and \JFnr as described in \Cref{sec:Metrics}.
Among them, \JFn is selected as the primary evaluation metric.

\myparagraph{Dataset Splits.}
These videos are split into {3,666} training, {433} validation, and {614} testing videos, for model training, daily evaluation, and competition period evaluation\footnote{The testing set is used for evaluation during the competition periods, such as \url{https://pvuw.github.io/} and \url{https://lsvos.github.io/}.}, respectively. An additional 311 videos, originally used as the validation set in MOSEv1, are temporarily retained for compatibility and may later serve as a local validation set when MOSEv2 becomes the standard.

\subsection{Semi-supervised Video Object Segmentation}
Semi-supervised (or semi-automatic, one-shot) VOS offers the target's mask, bounding box, or points on the first frame as reference for segmenting the entire video.

\noindent$\bullet$~\textbf{Mask-initialization.}
This is the most common and actively studied setting in VOS.
In \Cref{tab:svos}, we benchmark two groups of mask-initialization semi-supervised VOS methods on \ournewdataset{2}. 
The first group includes traditional VOS methods, typically built on ResNet-50. The second group comprises SAM2-based variants, covering both SAM2-B+ and SAM2-L scales.
Existing methods perform substantially worse on \ournewdataset{2} compared to previous benchmarks such as DAVIS$_{17}$~\cite{davis2017} and YouTube-VOS$_{19}$~\cite{youtube_vos}. For example, SAM2-B+~\cite{SAM2} achieves only 47.1\% \JF on \ournewdataset{2}, far below its 74.7\% \JF on \ournewdataset{1}, 83.1\% \JF on LVOSv2, and 90.2\% \JF on DAVIS$_{17}$.
Among traditional methods, Cutie-B~\cite{cutie} performs best with 42.8\% \JFn, just 3.2\% behind SAM2-B+. However, its strength mainly comes from high \JFnd scores (64.5\%) in disappearance handling, while struggling with reappearance, reaching just 18.3\% \JFnr, 4.9\% lower than SAM2-B+.

Taking a close look at the detailed metrics, we observe that all methods face significant challenges in reappearance scenarios, with \JFnr scores ranging only from 7.8\% to 34.9\%. This underscores the difficulty of re-identifying objects after disappearance in complex scenes. The proposed adaptive boundary metric \Fn consistently yields lower values than \F across all methods, showing that the adaptive threshold provides a stricter and more reliable assessment of boundary quality for objects of different sizes.

\begin{table}[t]
    \centering
    \renewcommand\arraystretch{1.12}
    \setlength{\tabcolsep}{1.56pt}
    \footnotesize
    \caption{Benchmarking box-initialization semi-supervised VOS methods on \ournewdataset{2} validation set.}
    \label{tab:box-svos}
    \vspace{-2.16mm}
    \resizebox{0.488\textwidth}{!}{
    \begin{tabular}{l|c|cccc|c|c}
    \thickhline
    \rowcolor{mygray3}       & &\multicolumn{4}{c|}{\textbf{\ournewdataset{2}}} & \ournewdataset{1} & DAVIS$_{17}$ \\
    \rowcolor{mygray3}\multirow{-2}{*}{Method}& \multirow{-2}{*}{Pub.}&\JFn & \JFnd & \JFnr & \JF &\JF & \JF \\ 
    \hline
    \hline
    UniVS~\cite{univs}& \pub{CVPR'24} & 16.4 & 22.3 & \ \ 8.6 & 17.3 & 38.0 & 61.8 \\
    Cutie+SAM   & \pub{CVPR'24} & 42.3 & 64.4 & 18.0 & 43.5 & 63.0 & 82.3 \\
    \hline
    SAM2-B+     & \pub{ICLR'25} & 46.0 & 61.9 & 22.1 & 47.2 & 73.7 & 85.3 \\
    SAMURAI-B+  & \pub{Preprint'24}   & 46.5 & 45.7 & 32.7 & 48.0 & 71.8 & 86.1 \\
    DAM4SAM-B+  & \pub{CVPR'25} & 46.2 & 49.9 & 31.3 & 47.6 & 70.1 & 86.6 \\
    SAM2Long-B+ & \pub{ICCV'25} & 47.7 & 57.4 & 28.3 & 49.0 & 72.9 & 85.5 \\
    \hline
    SAM2-L      & \pub{ICLR'25} & 49.0 & 61.9 & 26.2 & 50.3 & 75.4 & 89.0 \\
    SAMURAI-L   & \pub{Preprint'24}   & 49.2 & 49.9 & 33.8 & 50.7 & 74.9 & 88.9 \\
    DAM4SAM-L   & \pub{CVPR'25} & 47.5 & 51.5 & 32.2 & 48.8 & 73.3 & 86.6 \\
    SAM2Long-L  & \pub{ICCV'25} & 50.2 & 60.6 & 29.8 & 51.5 & 75.9 & 88.3 \\
    \thickhline
    \end{tabular}
    }
    \vspace{-2.16mm}
\end{table}

SAM2-based methods~\cite{SAM2} achieve superior performance, with even zero-shot models outperforming most finetuned traditional methods, demonstrating the effectiveness of foundation models on challenging video segmentation. Beyond SAM2, several SAM2-based variants~\cite{SAMURAI,dam4sam,SAM2Long} demonstrate enhanced performance on MOSEv2. These methods are specifically designed to address complex scenarios:
SAMURAI~\cite{SAMURAI} incorporates Kalman filtering for motion modeling to handle occlusions, DAM4SAM~\cite{dam4sam} introduces robust memory mechanisms to reduce distractor effects in crowded scenes, and SAM2Long~\cite{SAM2Long} employs a memory tree to mitigate error accumulation in long videos with object disappearance and reappearance. While these designs target specific challenges, they fall short on \ournewdataset{2}, where occlusions and long-term tracking are more severe. In addition, new challenges such as adverse environments, multi-shot transitions, and knowledge-dependent scenarios remain unaddressed. For example, SAM2Long-L~\cite{SAM2Long} achieves the best overall performance at only 51.5\% \JFn, highlighting substantial room for improvement in addressing complex real-world scenarios. Most SAM2 variants also show improved \JFnr but decreased \JFnd, reflecting a tendency toward aggressive re-identification. SAMURAI~\cite{SAMURAI} achieves the highest \JFnr but sacrifices the most in \JFnd, while SAM2Long provides a better balance and thus the best overall performance. The frequent disappearance–reappearance patterns and diverse complex scenarios in \ournewdataset{2} impose dual demands on both recall and precision. Future models must suppress false predictions when targets are absent yet reliably re-identify them upon reappearance. Effectively balancing these competing objectives remains a key challenge for future research.

\begin{table}[t]
    \centering
    \renewcommand\arraystretch{1.12}
    \setlength{\tabcolsep}{1.5pt}
    \footnotesize
    \caption{Benchmarking~point-initialization~semi-supervised VOS methods on \ournewdataset{2} validation set. We use \JFn as the evaluation metric. ``n-clk'': using n positive clicks for initialization.
    }
    \label{tab:point-svos}
    \vspace{-2mm}
    \resizebox{0.488\textwidth}{!}{
    \begin{tabular}{l|c|ccc|ccc|c}
    \thickhline
    \rowcolor{mygray3}   & &\multicolumn{3}{c|}{\textbf{\ournewdataset{2}}} & \multicolumn{3}{c|}{\ournewdataset{1}} & DAVIS$_{17}$ \\
    \rowcolor{mygray3}\multirow{-2}{*}{Method}& \multirow{-2}{*}{Pub.}& 1-clk & 3-clk & 5-clk & 1-clk & 3-clk & 5-clk & 5-clk \\ \hline
    \hline
    Cutie+SAM   & \pub{CVPR'24}  & 35.2    & 38.2    & 36.7    & 54.2    & 58.5    & 58.3    & 62.7    \\
    \hline
    SAM2-B+     & \pub{ICLR'25}  & 43.6    & 44.1    & 44.4    & 66.8    & 66.8    & 70.6    & 80.4    \\
    SAMURAI-B+  & \pub{Preprint'24} & 44.7    & 45.8    & 45.9    & 65.7    & 65.7    & 68.6    & 78.9    \\
    DAM4SAM-B+  & \pub{CVPR'25}  & 43.8    & 45.6    & 45.8    & 66.3    & 66.3    & 69.3    & 80.3    \\
    SAM2Long-B+ & \pub{ICCV'25}  & 45.3    & 45.3    & 45.1    & 66.4    & 66.4    & 70.3    & 80.5    \\
    \hline
    SAM2-L      & \pub{ICLR'25}  & 47.6    & 48.0    & 47.2    & 69.6    & 69.6    & 74.8    & 86.0    \\
    SAMURAI-L   & \pub{Preprint'24} & 47.9    & 48.2    & 48.6    & 69.3    & 69.3    & 74.1    & 84.8    \\
    DAM4SAM-L   & \pub{CVPR'25}  & 47.7    & 47.7    & 48.2    & 69.4    & 69.4    & 74.4    & 85.5    \\
    SAM2Long-L  & \pub{ICCV'25}  & 48.5    & 48.3    & 48.7    & 69.7    & 69.7    & 75.2    & 86.1    \\
    \thickhline
    \end{tabular}
    }
    \vspace{-2.16mm}
\end{table}

In terms of computational efficiency, there is a clear trade-off between accuracy and speed. Traditional methods such as XMem~\cite{xmem} and STCN~\cite{STCN} run faster at 49.8 and 45.1 FPS, respectively, but with lower performance. In contrast, SAM2-based methods achieve better results but at the cost of slower inference, with SAM2Long-L running at 7.1 FPS, and greater memory usage of 6.8 GiB compared to 0.9 GiB for Cutie-B~\cite{cutie}.

\noindent$\bullet$~\textbf{Box-initialization.} We benchmark box-initialization semi-supervised VOS methods on \ournewdataset{2} in \Cref{tab:box-svos}. Similar to the mask-initialization setting, we evaluate both traditional methods (UniVS~\cite{univs} and Cutie+SAM) and SAM2-based variants. The results show that SAM2-based methods clearly outperform traditional ones, with SAM2Long-L~\cite{SAM2Long} achieving the best performance of 50.2\% \JFn. However, all methods struggle with reappearance scenarios evaluated by \JFnr, while performing relatively better on disappearance cases evaluated by \JFnd. The performance gap between \ournewdataset{2} and other benchmarks such as DAVIS$_{17}$~\cite{davis2017} highlights the increased difficulty posed by the diverse and complex scenarios in \ournewdataset{2}.

\noindent$\bullet$~\textbf{Point-initialization.} 
We benchmark point-initialization semi-supervised VOS methods on \ournewdataset{2} in \Cref{tab:point-svos}, including Cutie+SAM and SAM2-based variants. The results show that SAM2-based methods significantly outperform Cutie+SAM, with SAM2Long-L achieving the best performance of 48.5\% \JFn using only a single click. However, increasing the number of clicks from 1 to 5 does not consistently improve results, and some methods even degrade. This sensitivity to point initialization suggests that ambiguity from point prompts, combined with the complex scenes in \ournewdataset{2}, makes it difficult for models to maintain stable segmentation despite additional user input.
Compared to DAVIS$_{17}$, where methods achieve much higher scores, \eg, 86.1\% \JFn for SAM2Long-L, the large performance gap further underscores the challenges of point-based setting in \ournewdataset{2}.

\begin{table}[t]
    \centering
    \renewcommand\arraystretch{1.12}
    \setlength{\tabcolsep}{1.2pt}
    \footnotesize
    \caption{Benchmark results of unsupervised VOS methods on \ournewdataset{2} validation set. We limit the number of proposals to 20.}
    \label{tab:zero-vos}
    \vspace{-2mm}
    \resizebox{0.488\textwidth}{!}{
    \begin{tabular}{l|c|cccc|c|c}
    \thickhline
    \rowcolor{mygray3}       & &\multicolumn{4}{c|}{\textbf{\ournewdataset{2}}} & \ournewdataset{1} & DAVIS$_{17}$ \\
    \rowcolor{mygray3}\multirow{-2}{*}{Method}& \multirow{-2}{*}{Pub.}&\JFn & \JFnd & \JFnr & \JF &\JF & \JF \\ 
    \hline
    \hline
    DEVA~\cite{deva}             & \pub{ICCV'23}    & 34.9 & 80.4 & \ \ 7.5  & 36.0 & 57.0 & 73.4 \\
    EntitySAM~\cite{entitysam}        & \pub{CVPR'25}    & 28.2 & 96.7 &\ \  4.1  & 28.4 & 42.2 & 72.6 \\
    \hline
    SAM2-B+          & \pub{ICLR'25}    & 28.3 & 77.3 & \ \ 6.3  & 28.8 & 47.2 & 57.3 \\
    SAMURAI-B+       & \pub{Preprint'24}& 27.5 & 52.5 & 12.0 & 28.4 & 46.9 & 57.4 \\
    DAM4SAM-B+       & \pub{CVPR'25}    & 25.9 & 52.5 & \ \ 6.9  & 26.7 & 46.4 & 57.7 \\
    SAM2Long-B+      & \pub{ICCV'25}    & 28.9 & 61.0 & \ \ 7.5  & 29.7 & 47.6 & 57.4 \\
    \hline
    SAM2-L           & \pub{ICLR'25}    & 28.2 & 73.5 & \ \ 6.3  & 28.6 & 48.1 & 57.9 \\
    SAMURAI-L        & \pub{Preprint'24}& 29.2 & 53.9 & 12.6 & 30.1 & 46.5 & 57.7 \\
    DAM4SAM-L        & \pub{CVPR'25}    & 28.7 & 52.6 & 10.6 & 29.6 & 47.8 & 58.0 \\
    SAM2Long-L       & \pub{ICCV'25}    & 29.1 & 52.6 & \ \ 8.8  & 29.8 & 48.3 & 58.0 \\
    \thickhline
    \end{tabular}
    }
\end{table}

\subsection{Unsupervised Video Object Segmentation}
Unsupervised (or automatic, zero-shot) VOS aims to automatically identify and segment primary objects in videos without manual guidance. Following DAVIS~\cite{davis2016,davis2017}, we limit the number of proposals to 20 for a fair comparison. In \Cref{tab:zero-vos}, we benchmark unsupervised VOS methods on \ournewdataset{2}. The results show that all methods perform poorly on \ournewdataset{2}, especially in reappearance cases, where \JFnr scores drop to as low as 4.1\%–12.6\%. Although DEVA~\cite{deva} achieves the highest \JF of 36.0\%, this remains far below its 73.4\% performance on DAVIS$_{17}$. For SAM2-based methods, we use grid prompts on the first frame to generate candidate masks, which are then propagated to subsequent frames. However, incomplete initial masks limit their effectiveness, with SAM2Long-L reaching only 29.1\% \JFn. The substantial performance gap between \ournewdataset{2} and other benchmarks highlights the challenging nature of our dataset for unsupervised VOS methods, which must handle complex scenes without any manual guidance.

\begin{table}[t]
    \centering
    \renewcommand\arraystretch{1.12}
    \setlength{\tabcolsep}{1.4pt}
    \footnotesize
    \caption{Benchmark results of interactive VOS methods on \ournewdataset{2} validation set.
    }
    \label{tab:ivos}
    \vspace{-2mm}
    \begin{tabular}{l|c|cc|c|c}
    \thickhline
    \rowcolor{mygray3}       & &\multicolumn{2}{c|}{\textbf{\ournewdataset{2}}} & \ournewdataset{1} & DAVIS$_{17}$ \\
    \rowcolor{mygray3}\multirow{-2}{*}{Method}& \multirow{-2}{*}{Pub.}& AUC \JF & $\mathcal{J}\&\mathcal{F}@60s$ & $\mathcal{J}\&\mathcal{F}@60s$ & $\mathcal{J}\&\mathcal{F}@60s$ \\ 
    \hline
    \hline
    MANet~\cite{MANet}&\pub{CVPR'20}& 28.9 & 41.2 & 46.0 &79.5\\
    CiVOS~\cite{civos}&\pub{CVPR'21}& 32.7 & 46.1 & 51.7 &84.0\\
    MiVOS~\cite{MiVOS}&\pub{CVPR'21}& 36.7 & 48.9 & 53.9 &88.5\\
    STCN~\cite{STCN} &\pub{NeurIPS'21}& 39.8 & 54.1 & 59.5 & 88.8\\
    \thickhline
    \end{tabular}
    \vspace{-1.6mm}
\end{table}

\begin{table}[t]
    \centering
    \renewcommand\arraystretch{1.12}
    \setlength{\tabcolsep}{1.1pt}
    \footnotesize
    \caption{Benchmark results of video object tracking (VOT) methods on \ournewdataset{2} validation set. AUC: area under the success curve; P and P$_\text{norm}$: precision metrics measuring center location error (raw and size-normalized); AO: average overlap.}
    \label{tab:vot}
    \vspace{-2mm}
    \resizebox{0.488\textwidth}{!}{
    \begin{tabular}{l|c|ccc|c|c|c}
    \thickhline
    \rowcolor{mygray3}   & &\multicolumn{3}{c|}{\textbf{MOSEv2}} & MOSEv1 & LaSOT & GOT-10k \\
    \rowcolor{mygray3}\multirow{-2}{*}{Method}& \multirow{-2}{*}{Pub.}& P & P$_\text{norm}$  & AUC & AUC & AUC & AO \\ 
    \hline
    \hline
    SeqTrack-B~\cite{seqtrack}  & \pub{CVPR'23}     & 21.3 & 24.8  & 23.7 & 42.9 & 71.5 & 74.5 \\
    AQATrack-B~\cite{aqatrack}  & \pub{CVPR'24}     & 22.6 & 25.6  & 24.6 & 44.7 & 72.7 & 76.0 \\
    ODTrack-B~\cite{odtrack}   & \pub{AAAI'24}     & 21.3 & 23.8  & 23.5 & 47.2 & 73.2 & 77.0 \\
    LORAT-B~\cite{lorat}     & \pub{ECCV'24}     & 20.8 & 24.1  & 23.3 & 43.8 & 71.7 & 72.1 \\
    SUTrack-B~\cite{sutrack}   & \pub{AAAI'25}     & 24.3 & 26.4  & 26.0 & 46.9 & 74.4 & 79.3 \\
    SAM2-B+     & \pub{ICLR'25}     & 29.2 & 30.0  & 29.1 & 58.3 & 66.0 & -    \\
    SAMURAI-B+  & \pub{Preprint'24} & 35.2 & 35.5  & 34.3 & 59.5 & 70.7 & 79.6 \\
    DAM4SAM-B+  & \pub{CVPR'25}     & 35.0 & 35.4  & 33.9 & 59.5 & -    & -    \\
    SAM2Long-B+ & \pub{ICCV'25}     & 32.0 & 32.6  & 31.4 & 58.3 & -    & -    \\
    \hline
    SeqTrack-L~\cite{seqtrack}  & \pub{CVPR'23}     & 23.5 & 26.3  & 25.3 & 45.7 & 72.5 & 74.8 \\
    ODTrack-L~\cite{odtrack}   & \pub{AAAI'24}     & 24.4 & 26.7  & 25.9 & 49.1 & 74.0 & 78.2 \\
    LORAT-L~\cite{lorat}     & \pub{ECCV'24}     & 23.6 & 26.7  & 25.5 & 46.0 & 75.1 & 77.5 \\
    SUTrack-L~\cite{sutrack}   & \pub{AAAI'25}     & 26.9 & 28.4  & 27.8 & 48.6 & 75.2 & 81.5 \\
    SAM2-L      & \pub{ICLR'25}     & 33.1 & 33.6  & 32.1 & 59.6 & 70.0 & 80.7 \\
    SAMURAI-L   & \pub{Preprint'24} & 37.4 & 37.8  & 36.1 & 60.9 & 74.2 & 81.7 \\
    DAM4SAM-L   & \pub{CVPR'25}     & 36.8 & 37.3  & 35.6 & 60.9 & 75.1 & -    \\
    SAM2Long-L  & \pub{ICCV'25}     & 34.2 & 34.8  & 33.1 & 60.2 & 73.9 & 81.1 \\
    \thickhline
    \end{tabular}
    }
    \vspace{-1.6mm}
\end{table}

\begin{table*}[t]
\centering
\setlength{\tabcolsep}{0.7pt}
\renewcommand{\JFn}{\fontsize{5.5}{10}\selectfont $\mathcal{J}\&\dot{\mathcal{F}}$\xspace}
\renewcommand{\JFnd}{\fontsize{5.5}{10}\selectfont $\mathcal{J}\&\dot{\mathcal{F}}_d$\xspace}
\renewcommand{\JFnr}{\fontsize{5.5}{10}\selectfont $\mathcal{J}\&\dot{\mathcal{F}}_r$\xspace}

\caption{Attribute-based performance analysis on \ournewdataset{2} validation set, with attribute definitions detailed in \cref{tab:challenge}. 
The overall metric represents the average value across all attributes. The best score in each metric is highlighted in \textbf{bold}.
}
\label{tab:attribute}
\vspace{-2mm}
\fontsize{6.8}{10}\selectfont
\begin{tabular}{l|ccc|ccc|ccc|ccc|ccc|ccc|ccc|ccc|ccc|ccc}
\thickhline
\rowcolor{mygray3} & \multicolumn{3}{c|}{Overall} & \multicolumn{3}{c|}{OCC} & \multicolumn{3}{c|}{DR} & \multicolumn{3}{c|}{CRO} & \multicolumn{3}{c|}{DV} & \multicolumn{3}{c|}{CE} & \multicolumn{3}{c|}{NC} & \multicolumn{3}{c|}{LD} & \multicolumn{3}{c|}{MS} & \multicolumn{3}{c}{KD} \\
\rowcolor{mygray3} \multirow{-2}{*}{Method} & 
\JFn & \JFnd & \JFnr & \JFn & \JFnd & \JFnr & \JFn & \JFnd & \JFnr & \JFn & \JFnd & \JFnr & \JFn & \JFnd & \JFnr & \JFn & \JFnd & \JFnr & \JFn & \JFnd & \JFnr & \JFn & \JFnd & \JFnr & \JFn & \JFnd & \JFnr & \JFn & \JFnd & \JFnr \\
\hline
\hline
XMem & 31.7 & 55.5 & 12.6 & 36.8 & 56.9 & 14.9 & 30.8 & 57.8 & 13.6 & 30.8 & 52.9 & \ \ 9.2 & 24.3 & 54.5 & \ \ 8.3 & 34.0 & 52.2 & 12.7 & 34.6 & 50.4 & 16.2 & 30.7 & 77.1 & 10.7 & 33.2 & \textbf{57.2} & 16.2 & 30.2 & 40.3 & 11.3 \\
Cutie-B & 35.8 & \textbf{61.9} & 15.7 & 43.4 & \textbf{64.7} & 18.4 & 35.7 & \textbf{59.8} & 17.3 & 36.8 & \textbf{60.1} & 14.2 & 26.8 & \textbf{67.0} & \ \ 9.6 & 42.0 & \textbf{68.7} & 15.2 & 39.9 & {55.1} & 20.7 & 35.4 & \textbf{81.8} & 13.5 & 31.8 & 51.2 & 19.6 & \textbf{30.5} & 48.3 & 12.7 \\
SAM2-B+ (ZS) & 36.8 & 56.0 & 17.0 & 43.5 & 59.7 & 20.8 & 38.6 & 53.4 & 23.5 & 36.6 & 52.5 & 14.3 & 28.5 & 49.9 & 10.5 & 49.2 & 67.3 & 24.4 & 37.3 & 52.1 & 17.5 & 40.5 & 65.9 & 18.5 & 30.1 & 45.9 & 14.7 & 26.9 & \textbf{57.5} & \ \ 9.0 \\
SAM2-B+ & 40.7 & 57.0 & 21.4 & 47.1 & 61.6 & 23.7 & 41.5 & 53.8 & 26.4 & 42.5 & 56.2 & 21.5 & 35.1 & 48.8 & 18.5 & 52.6 & 66.9 & 28.5 & 43.1 & \textbf{55.2} & 22.4 & 42.5 & 72.4 & 22.7 & 34.0 & 46.9 & 18.5 & 27.8 & 51.3 & \ \ 9.9 \\
SAMURAI-B+ & 42.6 & 40.7 & \textbf{30.1} & 48.9 & 46.6 & \textbf{33.9} & 44.2 & 38.7 & \textbf{33.4} & 41.0 & 40.6 & \textbf{26.1} & 37.5 & 32.5 & 29.6 & 55.6 & 52.7 & \textbf{39.8} & 43.6 & 39.1 & \textbf{32.5} & 51.6 & 50.8 & \textbf{39.8} & 32.7 & 28.4 & 20.1 & 28.2 & 37.1 & \textbf{15.4} \\
DAM4SAM-B+ & 42.4 & 46.4 & 27.9 & 48.7 & 52.1 & 32.2 & \textbf{44.5} & 47.2 & 31.5 & 40.9 & 45.5 & 24.7 & \textbf{39.4} & 43.2 & \textbf{29.7} & 52.9 & 56.6 & 32.8 & 43.8 & 44.5 & 30.6 & 51.1 & 56.9 & 35.5 & \textbf{34.1} & 31.5 & 19.3 & 25.9 & 40.4 & 14.5 \\
SAM2Long-B+ & \textbf{42.9} & 52.8 & 26.2 & \textbf{49.4} & 59.7 & 29.9 & 42.9 & 50.1 & 28.9 & \textbf{44.6} & 53.0 & 24.7 & 37.5 & 50.1 & 23.2 & \textbf{56.7} & 65.4 & 35.5 & \textbf{43.9} & 53.4 & 28.2 & \textbf{52.9} & 65.9 & 35.4 & 32.0 & 33.4 & \textbf{20.3} & 25.7 & 44.1 & \ \ 9.9 \\
\thickhline
\end{tabular}
\vspace{-1.6mm}
\end{table*}

\subsection{Interactive Video Object Segmentation}
Following the interactive track of the 2019 DAVIS Challenge on VOS~\cite{davis2019}, we provide initial scribbles for the target object as the first interaction. Interactive video object segmentation methods must predict the full-video segmentation based on this input. After comparing predictions with the ground truth, corrective scribbles are added on the worst-performing frame for refinement. This process can be repeated up to 8 times with a 30-second time limit per object. We report $\mathcal{J}\&\mathcal{F}$@60s to reflect the trade-off between accuracy and efficiency, measuring model performance within 60 seconds of interactive processing. As shown in \Cref{tab:ivos}, we evaluate four recent interactive VOS methods on \ournewdataset{2}. All methods show substantial performance drops compared to DAVIS$_{17}$. STCN~\cite{STCN} achieves the best performance of 54.1\% $\mathcal{J}\&\mathcal{F}$@60s, which is far below its 88.8\% on DAVIS$_{17}$. This significant performance gap highlights the increased difficulty of the complex scenarios in \ournewdataset{2}.

\subsection{Video Object Tracking}
Video object tracking (VOT) aims to track a target object throughout a video given an initial bounding box. Unlike VOS, VOT focuses on localization rather than segmentation. 
To adapt \ournewdataset{2} for VOT, we convert segmentation masks to bounding boxes by using the minimal enclosing rectangle.
In \Cref{tab:vot}, we benchmark nine state-of-the-art VOT methods on \ournewdataset{2}, including both traditional trackers and SAM2-based variants. 
Following LaSOT~\cite{lasot}, we adopt P, P$_\text{norm}$, and AUC as evaluation metrics.
The results show that all methods undergo a significant performance drop on \ournewdataset{2} compared to existing VOT benchmarks. 
Among traditional trackers, SUTrack-L~\cite{sutrack} performs best with 27.8\% AUC, while LORAT-B~\cite{lorat} performs worst with only 23.3\% AUC on \ournewdataset{2}. Overall, traditional methods remain weak, with scores between 23.3\% and 27.8\% AUC. SAM2-based methods achieve higher performance, with SAMURAI-L~\cite{SAMURAI} leading at 36.1\%, followed by DAM4SAM-L (35.6\%) and SAM2Long-L (33.1\%). However, these scores are far below those on other datasets. For example, SAMURAI-L achieves 60.9\% AUC on MOSEv1~\cite{MOSEv1}, 74.2\% on LaSOT~\cite{lasot}, and 81.7\% on GOT-10k~\cite{got-10k}, but drops to only 36.1\% on \ournewdataset{2}, underscoring the much greater challenges posed by \ournewdataset{2}. In addition, larger models (L variants) consistently outperform their base counterparts (B+ variants), suggesting that increased model capacity helps handle the diverse and challenging tracking conditions in \ournewdataset{2}.

Although SAMURAI~\cite{SAMURAI} underperforms other SAM2 variants such as SAM2Long~\cite{SAM2Long} in VOS tasks, it shows superior tracking performance in VOT. This is mainly due to two factors. First, VOT metrics do not penalize false positives when the ground truth is empty, which aligns with SAMURAI's higher \JFnr scores as shown in \Cref{tab:svos}. Second, the integration of Kalman filtering effectively captures temporal motion, enhancing localization and trajectory prediction in complex tracking scenarios.

\subsection{Attribute-Based Performance Analysis}

\begin{table}[t]
\centering
\renewcommand\arraystretch{1.1}
\setlength{\tabcolsep}{2.1pt}
\caption{Comparison on long videos (>300 frames) in MOSEv2 and LVOSv2. $\Delta$: the difference between \JFnd and \JFnr.}
\vspace{-2mm}
\label{tab:long_video}
\begin{tabular}{l|cccc|cccc}
\thickhline
 \rowcolor{mygray3}& \multicolumn{4}{c|}{\textbf{MOSEv2 (LD)}} & \multicolumn{4}{c}{LVOSv2} \\
 \rowcolor{mygray3} \multirow{-2}{*}{Method} & \JFn & \JFnd & \JFnr & $\Delta$ & \JFn & \JFnd & \JFnr & $\Delta$ \\
\hline
\hline
SAM2-B+    & 42.5 & 72.4 & 22.7 & +49.7 & 82.3 & 69.4 & 62.6 & +6.8  \\
SAMURAI-B+  & 51.6 & 50.8 & 39.8 & +11.0 & 81.5 & 56.8 & 71.3 & -14.5 \\
DAM4SAM-B+  & 51.1 & 56.9 & 35.5 & +21.4 & 81.4 & 65.7 & 71.4 & \ \ -5.7     \\
SAM2Long-B+ & 52.9 & 65.9 & 35.4 & +30.5 & 84.3 & 66.8 & 68.5 & \ \ -1.7 \\
\thickhline
\end{tabular}
\vspace{-2mm}
\end{table}

To better understand how different methods perform under specific challenges, \Cref{tab:attribute} presents an attribute-based analysis on \ournewdataset{2}. We evaluate mask-initialization semi-supervised VOS methods across nine representative attributes defined in \Cref{tab:challenge}, including occlusion (OCC), disappearance-reappearance (DR), crowding (CRO), diverse visibility (DV), complex environment (CE), novel categories (NC), long duration (LD), multi-shots (MS), and knowledge dependency (KD).

The results reveal several key insights about model performance across different challenges. 
1) SAM2Long-B+~\cite{SAM2Long} achieves the best overall performance with 42.9\% \JFn, consistent with its strong results in previous experiments, suggesting that robustness to \ournewdataset{2}'s challenges translates into better general effectiveness. 
2) Fine-tuning significantly improves SAM2-B+'s performance, raising its \JFn from 36.8\% to 40.7\%, which highlights the importance of adaptation to complex video scenarios. 
3) Traditional methods like Cutie-B~\cite{cutie} and XMem~\cite{xmem} excel in frames where objects are disappearing (\JFnd), with Cutie-B achieving the highest scores across most attributes (up to 81.8\% on LD). However, they struggle significantly on reappearance scenarios (\JFnr), often failing to re-identify targets. For example, Cutie-B scores 81.8\% \JFnd but only 13.5\% \JFnr on LD videos, indicating a tendency toward false negatives when objects reappear.
4) A comparison with LVOSv2~\cite{LVOSv2}, which specifically focuses on long videos, highlights that the long-duration sequences in \ournewdataset{2} involve not only extended frame counts but also greater scene complexity. As shown in \Cref{tab:long_video}, LVOSv2 exhibits small $\Delta$ values, \ie, the gap between \JFnd and \JFnr, indicating minimal difficulty in reappearance cases. In contrast, the LD subset of \ournewdataset{2} shows much larger $\Delta$ values (+11.0 to +49.7), indicating severe reappearance difficulty. These challenges arise from frequent occlusions, camera shot transitions, background clutter, ambiguous reappearance cases, \etc. For example, SAM2Long-B+ achieves 84.3\% \JFn on LVOSv2 but only 52.9\% on \ournewdataset{2}'s LD subset, underscoring the substantially more challenging nature of MOSEv2 dataset.
5) In knowledge-dependent (KD) scenarios, all methods demonstrate significantly degraded performance, with Cutie-B~\cite{cutie} achieving only 30.5\% \JFn, underscoring the complexity of KD challenges.
Traditional methods such as Cutie-B and XMem outperform SAM2 variants in KD scenarios, likely because they incorporate instance-level memory mechanisms that offer stronger semantic representation. SAM2~\cite{SAM2}, in contrast, is not pretrained on such scenarios and lacks heuristic design for knowledge-intensive tasks.
Among SAM2-based methods, SAMURAI-B+~\cite{SAMURAI} performs best in KD scenarios (15.4\% \JFnr), possibly due to its Kalman filter-based motion modeling provides spatial reasoning that is beneficial in certain KD cases requiring spatial cues.

\begin{figure*}[t!]
\centering 
\includegraphics[width=0.996\textwidth]{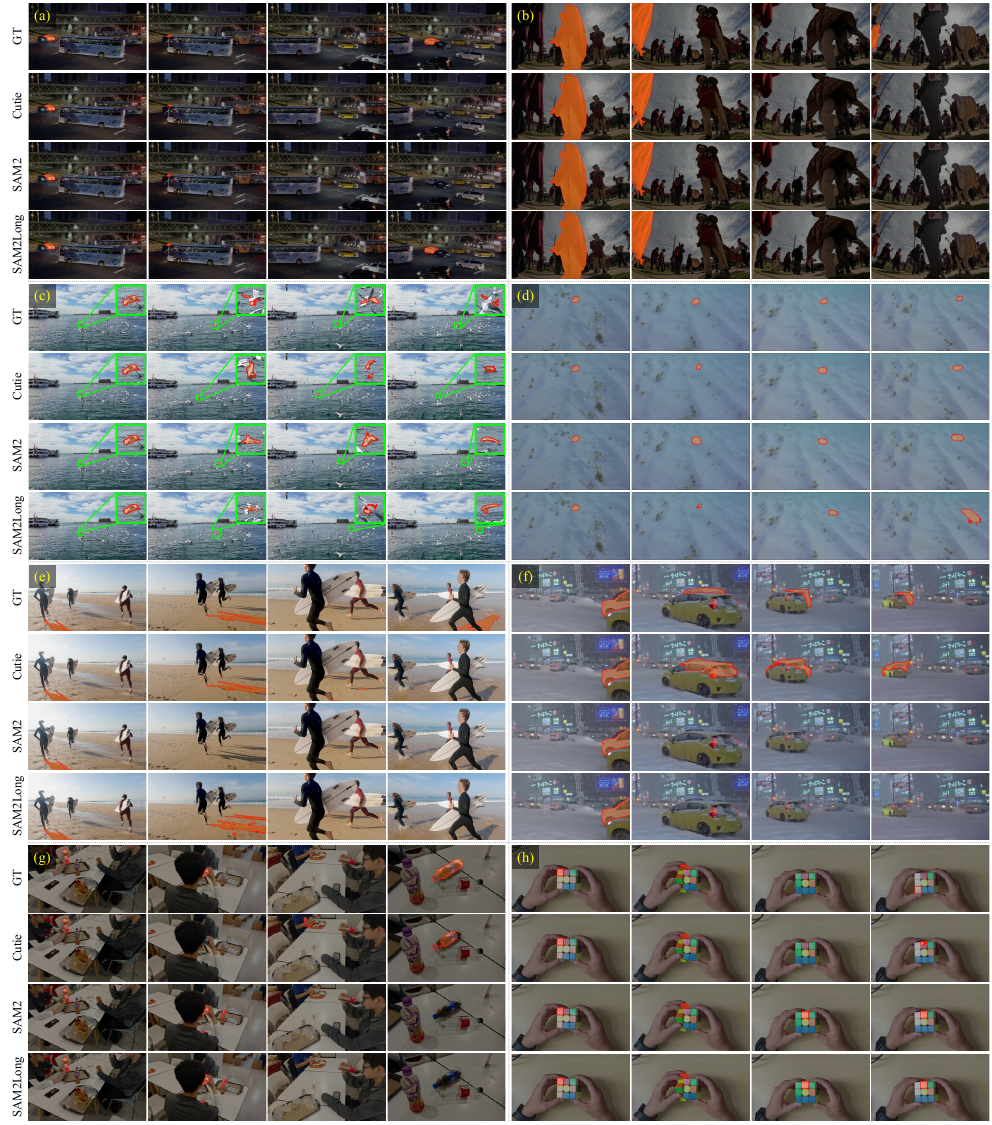} 
\vspace{-7.6mm}
\caption{
Qualitative results on \ournewdataset{2}. We compare Cutie~\cite{cutie}, SAM2~\cite{SAM2}, and SAM2Long~\cite{SAM2Long} on 8 challenging cases that assess model performance under various complex conditions. These include object disappearance and reappearance (a, b, e, g, h), small/inconspicuous objects (c), heavy occlusions (c, f), crowded scenes (c), adverse weather (f), low-light environments (a, d), multi-shot sequences (g), camouflaged targets (d), non-physical targets (e), and knowledge-dependent scenarios (h).}
\label{Fig:visualize}
\vspace{-2mm}
\end{figure*}
\subsection{Qualitative Analysis}

\cref{Fig:visualize} presents eight challenging cases that reveal key limitations of existing VOS methods. 1) Models struggle with re-identifying objects after disappearance and occlusion. While SAM2Long~\cite{SAM2Long}, which maintains multiple segmentation paths, successfully tracks a car undergoing simple linear motion (case a), it fails in more complex motion patterns such as a person walking around a crowd before reappearing (case b), indicating limitations in modeling long-term and nonlinear trajectories. 2) Densely crowded scenes containing small and heavily occluded targets (case c) remain extremely challenging, none of existing models succeed under such complexity. 3) In cases involving camouflaged objects or non-physical targets like shadows (cases d and e), Cutie~\cite{cutie} outperforms SAM2~\cite{SAM2} and SAM2Long~\cite{SAM2Long}, especially in boundary quality. This advantage may stem from Cutie’s compact instance-level memory, which explicitly models foreground objects and enables better separation from background distractions, while SAM2 relies on global image features lacking instance-specific cues. 4) Under adverse environmental conditions such as heavy snow (case f), the combination of low contrast and occlusion causes all models to fail, with Cutie producing inaccurate masks and SAM2 variants completely losing the target. 5) When faced with dramatic changes in viewpoint and object pose across multiple camera shots (case g), all models fail to maintain consistent tracking, as exemplified by the shifting appearance of a Coke bottle. 6) In scenarios that require understanding physical object relationships and transformation rules (case h), such as tracking a rotating Rubik's cube, the models fail to re-identify the correct block after disappearance, often incorrectly assigning it to adjacent blocks.

\begin{algorithm}[t]
    \caption{\small{Reliable Conditioned Memory Selection (RCMS).}}
    \label{alg:rcms_code}
    \definecolor{softgreen}{rgb}{0.1, 0.5, 0.1}
    \definecolor{codeblue}{rgb}{0.25,0.5,0.5}
    \definecolor{mygray2}{rgb}{0.9,0.9,0.9}
    \definecolor{oiBlue}{RGB}{0,114,178}
    \definecolor{oiOrange}{RGB}{230,159,0}
    \definecolor{oiSky}{RGB}{86,180,233}
    \definecolor{oiGreen}{RGB}{0,158,115}
    \definecolor{oiYellow}{RGB}{240,228,66}
    \definecolor{oiVermilion}{RGB}{213,94,0}
    \definecolor{oiPurple}{RGB}{204,121,167}
    
    \lstdefinelanguage{MyPython}{
        language=Python,
        morekeywords={[2]RCMS},
        alsoletter={_,<,>},
        morekeywords={[3]torch,cat,view},
        deletekeywords={shape},
    }
    \lstset{
        backgroundcolor=\color{white},
        basicstyle=\fontsize{6.5pt}{6.5pt}\ttfamily\selectfont,
        columns=fullflexible,
        breaklines=true,
        captionpos=b,
        commentstyle=\fontsize{6.5pt}{6.5pt}\color{softgreen},
        keywordstyle=\fontsize{6.5pt}{6.5pt},
        numbers=left,
        numberstyle=\tiny\color{gray},
        numbersep=-5pt,
        xleftmargin=0em,
    }
    \vspace{-0.075in}
    \begin{lstlisting}[language=MyPython]
    def RCMS(sam2_model, mask_0, frame_list, theta, N, K):
        # sam2_model: SAM2 model for video segmentation
        # M_0: Initial mask from the first frame
        # frame_list: List of video frames to process
        # K: Number of unconditioned memories, default 6
        # N: Maximum number of conditioned memories to select
        # theta: Quality threshold for memory selection
        # Process first frame to get initial memory
        memory_0 = sam2_model.init_state(frame_list[0], mask_0)
        cond_memory = [memory_0]
        all_memory = [] # Other memories expect cond
        masks = [mask_0]
        # Process remaining frames
        disappeared = False
        for t, frame in enumerate(frame_list[1:]):
            # Track object in current frame
            uncond_memory = select_nearest(all_memory, K)
            mask_t, memory_t = 
                sam2_model.step(frame, cond_memory, uncond_memory)
            all_memory.append(memory_t)
            masks.append(mask_t)
    \end{lstlisting}
    \vspace{-0.066in}
    {\lstset{
  basicstyle=\fontsize{6.5pt}{6.5pt}\ttfamily\bfseries\color{black!85}\selectfont,
  backgroundcolor=\color{black!6},
  aboveskip=0pt,belowskip=0pt}
    \begin{lstlisting}[language=MyPython, firstnumber=last]
            # Apply RCMS if object disappeared
            if is_empty(mask_t) and not disappeared:
                disappeared = True
                for i in range(len(all_memory) - 2, -1, -1): 
                    memory = all_memory[i]
                    Q = compute_quality_scores(memory)
                    if Q > theta and len(cond_memory) < N + 1:
                        cond_memory.append(memory)
                        all_memory.pop(i) 
    \end{lstlisting}}
    {\lstset{aboveskip=0pt,belowskip=0pt}
\begin{lstlisting}[language=MyPython, firstnumber=last]
        return masks
\end{lstlisting}}
\end{algorithm}

\begin{table}[t]
\centering
\renewcommand\arraystretch{1.12}
\setlength{\tabcolsep}{1.4pt}
\footnotesize
\caption{Ablation study of SAM2 improvements. * denotes SAM2-L with all improvements applied.}
\vspace{-3mm}
\label{tab:tricks}
\begin{tabular}{l|cccccccc|c}
\thickhline
\rowcolor{mygray3} Method              & \JFn & \J   & \Fn  & \JFnd & \JFnr & \F & \JF & \JFnbd &  FPS \\
\hline
\hline
SAMURAI-B+             & 47.4 & 45.3 & 49.5 & 45.9 & 33.6 & 52.2 & 48.8 & 73.3 & 17.7 \\
DAM4SAM-B+             & 47.9 & 45.8 & 50.0 & 51.3 & 32.0 & 52.6 & 49.2 & 73.1 & 17.3 \\
SAM2Long-B+            & 48.6 & 46.7 & 50.5 & 58.4 & 29.2 & 52.8 & 49.7 & 72.9 & \ \ 9.4  \\
\hline
SAM2-B+       & 46.0 & 44.2 & 47.8 & 61.6 & 23.2 & 50.0 & 47.1 & 73.5 & 23.4 \\
\rowcolor{mygray2}+RCMS w/o MQF         & 49.3 & 47.4 & 51.2 & 61.0 & 29.7 & 53.5 & 50.4 & 73.7 &  22.4 \\
\rowcolor{mygray2}\ \ \ \  +MQF     & 50.2 & 48.2 & 52.2 & 59.5 & 31.4 & 54.6 & 51.4 & 74.0 &  22.6 \\
\rowcolor{mygray2}\ \ \ \ \ \ \ \ +MSS         & 50.6 & 48.5 & 52.7 & 55.8 & 33.7 & 55.4 & 51.9 & 73.9 &  22.6 \\
\rowcolor{mygray2}\ \ \ \ \ \ \ \ \ \ \ \ +LVT         & 51.5 & 49.5 & 53.6 & 56.6 & 36.5 & 56.3 & 52.9 & 74.0 &  22.6 \\
\hline
SAM2-L & 49.7 & 47.9 & 51.5 & 64.5 & 27.1 & 53.8 & 50.9 & 74.6 & 14.4 \\
\rowcolor{mygray2} SAM2-L* (ours) & 54.4 & 52.4 & 56.3 & 66.8 & 33.2 & 58.9 & 55.6 & 75.6 & 14.3 \\
\thickhline
\end{tabular}
\vspace{-2.86mm}
\end{table}
\subsection{Enhancing SAM2 for Complex Scenarios}
\label{sec:trick}

Based on the above experimental results and the characteristics of \ournewdataset{2}, we introduce several practical improvements to SAM2.

\noindent$\bullet$~\textbf{Revisiting Memory Control in SAM2.}
SAM2 employs two types of memory: \textit{conditioned} and \textit{unconditioned}. The conditioned memory, typically derived from the initial frame, stores reliable object features that serve as strong references for object tracking and are particularly important for re-identification during disappearance–reappearance scenarios. The unconditioned memories, collected from the nearest temporal frames (up to 6), capture short-term appearance variations and motion dynamics. 
By default, however, the conditioned memory contains only a single frame from the initialization, which poses two major limitations. First, relying on a single frame restricts the diversity of object representations and limits the model’s ability to capture appearance variations. Second, if the initial frame provides only partial visibility of the target object (\eg, case \raisebox{-0.1em}{\scalebox{1.06}{\ding{173}}} in \cref{Fig:teaser}), the resulting memory lacks sufficient information for reliable segmentation in subsequent frames. This raises a key question: \textit{how can we obtain more reliable conditioned memories without incurring additional cost?}

\noindent$\bullet$~\textbf{Reliable Conditioned Memory Selection.} As shown in \Cref{tab:tricks}, SAM2-B+ achieves a \JFnbd of 73.5\%, a metric that evaluates performance exclusively during the initial continuous segment before the first disappearance of the target object. This indicates that SAM2 is highly robust in tracking continuously visible targets. 
Motivated by this observation, we propose Reliable Conditioned Memory Selection (RCMS), which preserves SAM2’s strong tracking ability in pre-disappearance frames while strategically augmenting the conditioned memory when disappearance occurs. As shown in \Cref{alg:rcms_code} (L22-30), RCMS selects the $N$ nearest high-quality memories from the pre-disappearance sequence and incorporates them into the conditioned memory bank.
To ensure memory reliability, we adopt Memory Quality Filtering (MQF). Specifically, for each candidate memory, a quality score is computed as: $Q = score_{iou} \times score_{occ} \times maskness$, where $score_{iou}$ and $score_{occ}$ are outputs of SAM2’s mask decoder indicating predicted IoU and occlusion confidence, respectively, and $maskness$~\cite{solo} measures mask quality. Only frames with $Q$ above a threshold $\theta$ are incorporated into the conditioned memory. In total, RCMS produces at most $N+1$ conditioned memories.

\begin{figure}[t!]
    \centering
    \includegraphics[height=2.5cm]{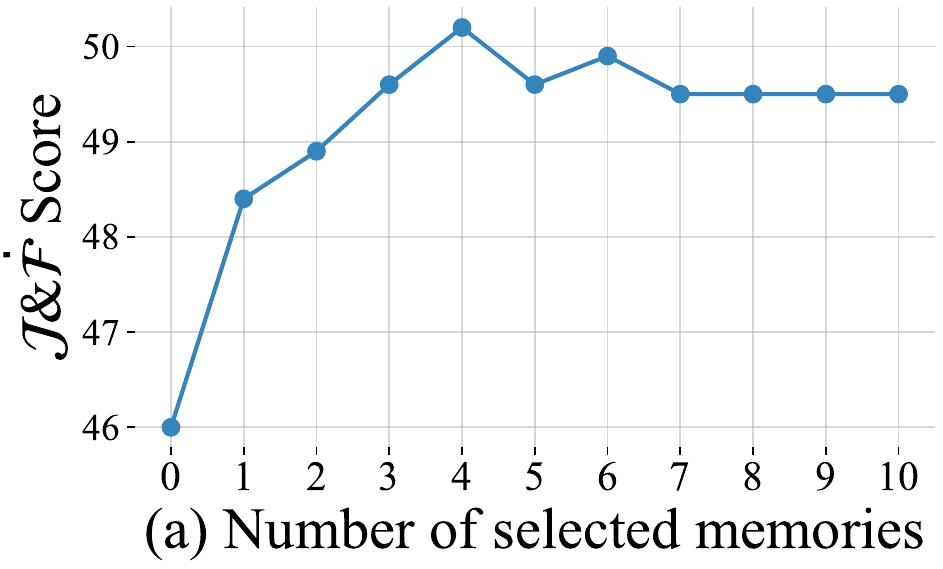}
    \includegraphics[height=2.5cm]{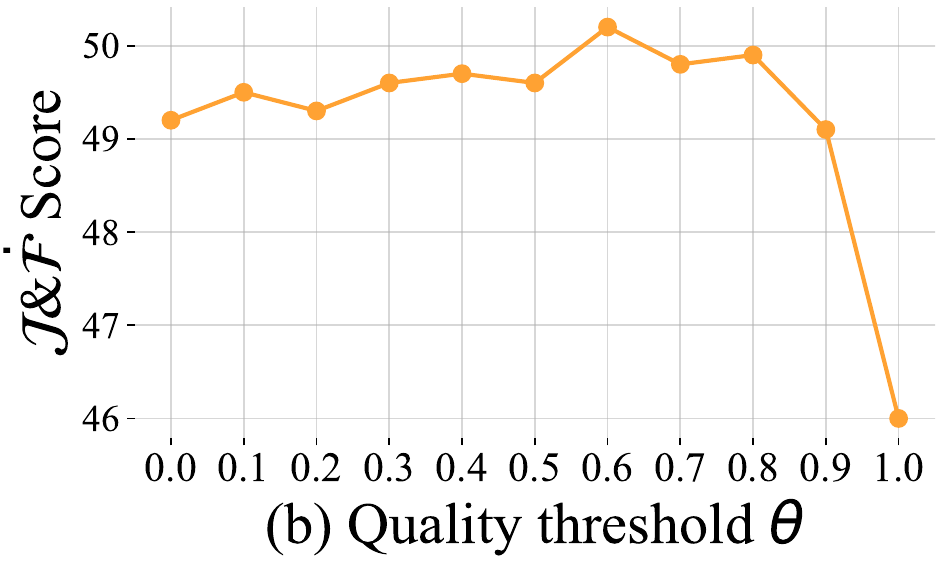}
    \vspace{-2mm}
    \caption{Ablation on RCMS and MQF. (a) number of selected conditioned memories. (b) quality threshold $\theta$ for memory selection.}
    \label{fig:rcms}
    \vspace{-2mm}
\end{figure}

As shown in \Cref{tab:tricks}, RCMS improves \JFn by +3.3\% (from 46.0\% to 49.3\%), with particularly strong gains in \JFnr (+6.5\%), confirming its effectiveness for object re-identification after disappearance. Incorporating MQF yields an additional +0.9\% improvement in \JFn, highlighting the importance of filtering for high-quality memories. \cref{fig:rcms} presents ablation studies on RCMS parameters. Performance increases steadily from $N=0$ (46.0\% \JFn) to $N=4$ (50.2\% \JFn), then saturates, suggesting that 4 additional memories provide sufficient diversity. For the quality threshold $\theta$, performance peaks at $\theta=0.6$ (50.2\% \JFn).
Compared to existing SAM2 variants, RCMS demonstrates clear advantages by leveraging reliable pre-disappearance memories.
DAM4SAM~\cite{dam4sam} also adds additional memories, but relies solely on threshold-based selection without considering timing. In contrast, RCMS leverages SAM2's strength in tracking continuously visible objects to introduce more reliable memories at appropriate moments. SAM2Long~\cite{SAM2Long} maintains multiple segmentation paths but suffers heavy computational overhead (9.4 FPS \vs our 22.6 FPS). Our method overcomes the limitation of depending solely on the initial frame memory while preserving efficiency and exploiting optimal timing for memory augmentation.

In addition, we incorporate the Mask Scaling Strategy (MSS) and Long-Video Finetuning (LVT) to further enhance SAM2. MSS adjusts mask output distributions with a scaling factor of 7.5 and an offset of –4.0~\cite{ding2025pvuw}, improving robustness to small objects and occlusions. LVT adapts SAM2 to challenging long-duration scenarios, which are particularly prevalent in \ournewdataset{2}. After the default finetuning with 8 frames, we perform additional finetuning with 16 frames while freezing the image encoder, enabling the model to better capture long-term temporal dependencies. As shown in \Cref{tab:tricks}, MSS yields a +0.4\% gain in \JFn with a notable +2.3\% improvement in \JFnr, while LVT further boosts performance by +0.9\% in \JFn and +2.8\% in \JFnr.

Together, these improvements raise SAM2-B+ from 46.0\% to 51.5\% \JFn, a substantial +5.5\% gain while retaining competitive inference speed. Applying all improvements to SAM2-L, denoted as SAM2-L* in \Cref{tab:tricks}, further boosts performance from 49.7\% to 54.4\% \JFn, a +4.7\% increase without sacrificing speed.
It is worth noting that SAM2-L* achieves lower \JFnr than SAM2-B+. Since the improvements primarily target re-identification after disappearance, they raise \JFnr for both backbones. However, the weaker B+ backbone struggles to balance \JFnd and \JFnr, often gaining higher \JFnr at the cost of lower \JFnd. In contrast, the stronger L backbone maintains a better balance, keeping both metrics at desirable levels.
\section{Discussion and Future Directions}\label{sec:Discussion}
Based on the comprehensive analysis of \ournewdataset{2} and the results of existing methods, we identify several key challenges and future research directions for complex video object segmentation.

\vspace{0.16mm}
\noindent$\bullet$\!~\textbf{Robust Re-identification for Disappearance-Reappearance.} 
The large drop in \JFnr scores reveals a key challenge in handling disappearance-reappearance, especially with complex motions, viewpoint changes, or knowledge-dependent cues (\eg, \cref{Fig:teaser}\raisebox{-0.1em}{\scalebox{1.06}{\ding{181}}}). Overly aggressive matching can inflate false positives during disappearance, reducing \JFnd. Future research should develop adaptive re-identification strategies that integrate appearance, motion, and semantic reasoning to better handle these scenarios.

\vspace{0.16mm}
\noindent$\bullet$\!~\textbf{Occlusion Handling.}
\ournewdataset{2} contains frequent and complex occlusions. Current methods often fail when objects are partly or fully hidden. Future work should design occlusion-aware models, such as attention for hidden regions, multi-scale feature fusion, and temporal models that keep object identity through occlusion.

\vspace{0.16mm}
\noindent$\bullet$\!~\textbf{Tracking in Crowded and Small-Target Scenarios.}
Small objects and crowded scenes often co-occur in \ournewdataset{2}, posing great challenges. Limited input resolutions (\eg, 480p in Cutie, 1024p in SAM2) lose fine details, making small-object tracking difficult. Future work should develop efficient ways to process high-resolution inputs and strengthen feature learning for small targets, such as multi-scale architectures, small-object–focused attention, and contrastive learning to distinguish targets from similar distractors in crowded scenes.

\vspace{0.16mm}
\noindent$\bullet$~\textbf{Generalization to Rare Categories.}
Although VOS methods are designed to be class-agnostic, generalizing to rare or unseen categories remains difficult. \ournewdataset{2} contains \numclass categories with a clear long-tail distribution, including uncommon targets like shadows and camouflaged objects. Current methods often fail on these categories due to limited training data and domain gaps. Future work could explore test-time adaptation using first-frame cues, or design stronger instance-level representations that better generalize to rare and visually ambiguous objects.

\vspace{0.16mm}
\noindent$\bullet$\!~\textbf{Environmental Robustness.}
\ournewdataset{2} covers diverse adverse environments, \eg, rain, snow, fog, nighttime, and underwater scenes, which severely degrade the performance of current VOS methods. In such settings, low visibility makes object appearance unreliable, while illumination changes and environmental occlusions disrupt temporal consistency. Future research should explore adaptive techniques such as weather-invariant and illumination-robust representations, as well as the integration of auxiliary signals or priors to improve robustness in real-world scenarios.

\vspace{0.16mm}
\noindent$\bullet$\!~\textbf{Multi-Shot Video Handling.}
Most methods rely on appearance matching and position estimation under temporal continuity, which fails in multi-shot videos where scene transitions cause abrupt changes in object appearance and position. Such structures are common in real-world content, especially narrative-driven videos. Future research should explore shot-aware strategies that handle discontinuities while preserving object identity across shots.

\vspace{0.16mm}
\noindent$\bullet$\!~\textbf{Knowledge-Dependent Tracking.}
Although recent methods have made progress in many VOS scenarios, they still struggle in cases requiring external knowledge such as spatial reasoning or common sense. These limitations stem from the fact that most models mainly rely on appearance and positional cues with limited reasoning ability. Future work could explore integrating MLLMs \cite{llava,internvl} to enhance semantic understanding and high-level reasoning. The main challenge is achieving this integration while preserving computational efficiency and real-time performance.

\section{Conclusion}
In this work, we introduce \ournewdataset{2}, a significantly more challenging dataset for video object segmentation in complex scenes. It extends \ournewdataset{1} in both scale and complexity of scenarios, comprising \numvideo high-resolution videos and \nummask object masks across \numclass categories. The dataset not only intensifies challenges in \ournewdataset{1}, such as object disappearance-reappearance, occlusions, and crowded scenes, but also introduces new challenges, such as adverse weather, low-light scenes, multi-shot sequences, camouflaged targets, non-physical targets, and knowledge-dependent cases. Evaluation across multiple VOS and VOT settings reveals that current state-of-the-art methods suffer significant performance drops on MOSEv2. For example, SAM2 drops from 90.7\% \JF on DAVIS 2017 to 50.9\% on \ournewdataset{2}. These results highlight the gap between existing algorithms and the demands of real-world deployment. Based on the analysis of the observed challenges, several practical tricks are proposed, which substantially enhance model performance with a 5.5\% \JFn gain. We believe \ournewdataset{2} will serve as a valuable resource for advancing robust and generalizable video object segmentation and tracking in diverse and unconstrained environments.

{
\bibliographystyle{IEEEtran}
\bibliography{egbib}
}

\end{document}